\documentclass[letterpaper,twocolumn,10pt]{article}

\usepackage{multirow}
\usepackage{usenix,epsfig,endnotes}

\usepackage{caption}

\usepackage{xspace}
\usepackage{etoolbox}
\usepackage{times}
\usepackage{adjustbox}
\usepackage{caption}
\usepackage{subcaption}
\usepackage[table]{xcolor}
\usepackage{xcolor}
\usepackage{amsmath}
\usepackage{cleveref}
\usepackage{flushend}
\usepackage{hyperref}
\usepackage[ruled,vlined]{algorithm2e}
\usepackage{algorithmic}
\usepackage[normalem]{ulem}
\usepackage{booktabs}
\usepackage{wrapfig}
\usepackage{flushend}
\usepackage{enumitem}

\newcommand{\myparagraph}[1]{\noindent\textbf{#1. }}

\newcommand{\system}{DiffKV\xspace}

\newcommand{\llama}{Llama}

\newtoggle{showmarks}
\toggletrue{showmarks}  

\iftoggle{showmarks}{
  \newcommand\inigo[1]{\textcolor{red}{IG: #1}}
  \newcommand\gohar[1]{\textcolor{orange}{GOHAR: #1}}
  \newcommand\rodrigo[1]{\textcolor{cyan}{RF: #1}}
  \newcommand\ricardo[1]{\textcolor{blue}{RB: #1}}
  \newcommand\sameh[1]{\textcolor{brown}{SE: #1}}
  \newcommand\yanqi[1]{\textcolor{cyan}{YANQI: #1}}
  \newcommand\yuwei[1]{\textcolor{orange}{YUWEI: #1}}
  \newcommand\TODO[1]{\textcolor{red}{TODO: #1}}
  \newcommand\CR[1]{\textcolor{blue}{#1}}
}{
  \newcommand\inigo[1]{\unskip}
  \newcommand\gohar[1]{\unskip}
  \newcommand\rodrigo[1]{\unskip}
  \newcommand\ricardo[1]{\unskip}
  \newcommand\sameh[1]{\unskip}
  \newcommand\yanqi[1]{\unskip}
  \newcommand\yuwei[1]{\unskip}
  \newcommand\TODO[1]{\unskip}
  \newcommand\CR[1]{\unskip}
 }

\begin{document}


\title{\textit{\system}: Differentiated Memory Management for Large Language Models with Parallel KV Compaction}

\makeatletter
\def\@fnsymbol#1{}
\makeatother

\author{
    \textnormal{Yanqi Zhang\textsuperscript{1}\textsuperscript{*}}\hspace{0.5em}
    \textnormal{Yuwei Hu\textsuperscript{1}\textsuperscript{*}}\hspace{0.5em}
    \textnormal{Runyuan Zhao\textsuperscript{1}}\hspace{0.5em}
    \textnormal{John C.S. Lui\textsuperscript{2}}\hspace{0.5em}
    \textnormal{Haibo Chen\textsuperscript{3}\textsuperscript{$\dagger$}}
    \\
    \textnormal{\textsuperscript{1}Huawei}\hspace{1em}
    \textnormal{\textsuperscript{2}The Chinese University of Hong Kong}\hspace{1em}
    \textnormal{\textsuperscript{3}Shanghai Jiao Tong University}
    \\
    \thanks{*Both authors contributed equally to this research.}
    \thanks{$\dagger$Haibo Chen is the corresponding author.}
}

\maketitle


\abstract{

Large language models (LLMs) demonstrate remarkable capabilities but face substantial serving costs due to their high memory demands, with the key-value (KV) cache being a primary bottleneck. State-of-the-art KV cache compression techniques, such as quantization and pruning, apply uniform treatment to both keys and values, and discard unimportant tokens entirely, overlooking the fine-grained distinctions in the significance of individual KV cache components.
To address such limitations, we introduce \textit{\system}, a novel framework for efficient KV cache compression that exploits three levels of differentiation in the KV cache: (1) the differing impact of keys and values on attention computation, (2) the varying importance of tokens, and (3) the diverse dynamic sparsity patterns across attention heads.
These levels of differentiation introduce irregular memory usage patterns across different requests and attention heads, posing significant scalability challenges for memory management. To address these challenges, \system proposes an on-GPU memory manager that compacts fragmented free memory list into contiguous regions in parallel, effectively translating sparsity in the KV cache into performance gains.
We evaluate \system on several mainstream LLMs, including the emerging thinking models that generate extended chains of thought. \system is able to compress the KV cache by $2.7\times$ to $5.7\times$ with near-lossless accuracy on complex workloads requiring sophisticated reasoning and long-generation capabilities, and enhances throughput by $1.9\times$ to $5.4\times$.
Source codes of \system are available at \url{https://github.com/zyqCSL/DiffKV}.
}
\vspace{-1em}
\section{Introduction}
\label{sec:intro}
\vspace{-1.2em}


Large language models (LLMs) like GPT~\cite{brown2020language,achiam2023gpt, ouyang2022training} and Gemini~\cite{team2023gemini} have demonstrated significant potential to impact our daily lives, offering promising applications in areas including chatbots~\cite{vicuna,chiang2024chatbot}, programming~\cite{gitcopilot, guo2024deepseek}, mathematics~\cite{shao2024deepseekmath, zhong2024evaluation} and medical assistance~\cite{saab2024capabilities, xie2024preliminary}. 
Despite their exceptional performance, hosting LLMs is costly due to their large model sizes, demanding extensive hardware resources. Given the pervasive adoption of LLMs, enhancing serving efficiency has become critically important~\cite{aminabadi2022deepspeed, fang2021turbotransformers, kao2023flat, shi2023welder, zhang2023shepherd, han2022microsecond, pope2023efficiently, li2023alpaserve}.

To avoid redundant computation, LLM inference frameworks typically cache intermediate key and value tensors in memory, commonly referred to as the KV cache~\cite{kwon2023efficient}. The size of the KV cache scales linearly with both sequence length and the number of concurrent requests, often comprising over 90\% of total memory consumption~\cite{kwon2023efficient,yu2022orca,gao2018low,yang2022infless}. 
As state-of-the-art models continue to support longer sequences~\cite{dubey2024llama,jiang2024mixtral,liu2024deepseek,achiam2023gpt,team2023gemini,touvron2023llama}, and with the rise of recent thinking models~\cite{guo2025deepseek, jaech2024openai, qwq32b} that generate extended reasoning processes, the KV cache has emerged as a critical bottleneck for LLM serving efficiency. 
It limits the number of concurrent requests and increases attention computation latency due to its memory bandwidth bound nature~\cite{dao2022flashattention,dao2023flashattention}.


Researchers have investigated various compression techniques, primarily focused on pruning~\cite{zhang2024h2o, xiao2023efficient, zhang2024pyramidkv, ge2023model, li2024snapkv} and quantization~\cite{lin2024qserve, flashinfer, hooper2024kvquant}. 
Pruning reduces the token sequence length by eliminating unimportant tokens based on attention scores.
Quantization, on the other hand, reduces the size of the KV cache by converting floating-point feature values into lower-precision representations.

While effective, these methods overlook the nuanced variations in importance across different components of the KV cache. 
First, existing pruning approaches employ a one-size-fits-all strategy, statically allocating memory uniformly across attention heads despite the per-head dynamic attention sparsity patterns.
Second, current quantization schemes apply uniform precision to all key and value vectors, ignoring both the distinct roles of keys versus values in the attention mechanism, and the varying significance of different tokens.
Finally, both pruning and quantization treat all requests uniformly, overlooking variability in information density across requests.




To address these limitations and advance KV cache compression, we propose \textit{\system}, a novel framework that exploits three levels of differentiation within the KV cache:

\myparagraph{1. Differentiated Keys and Values}
\system assigns higher precision to key vectors than to value vectors, based on the observation that keys (critical for attention score computation) play a more significant role than values (used primarily for weighted aggregation) in attention computations.

\myparagraph{2. Token Importance Differentiation}
Guided by attention scores, \system stores tokens at varying precision levels based on their significance. Most critical tokens are quantized at higher precision, less critical tokens are quantized at lower precision, and the least significant tokens are pruned. 

\myparagraph{3. Per-head Dynamic Sparsity}
Attention score distributions exhibit dynamic sparsity across both heads and requests: the number of important tokens varies across different heads, and for the same head under different requests.
To address this variability, \system dynamically identifies critical tokens per-head, per-request, adapting memory allocation accordingly.

\begin{figure}[t]
\centering
\includegraphics[width=0.95\linewidth]{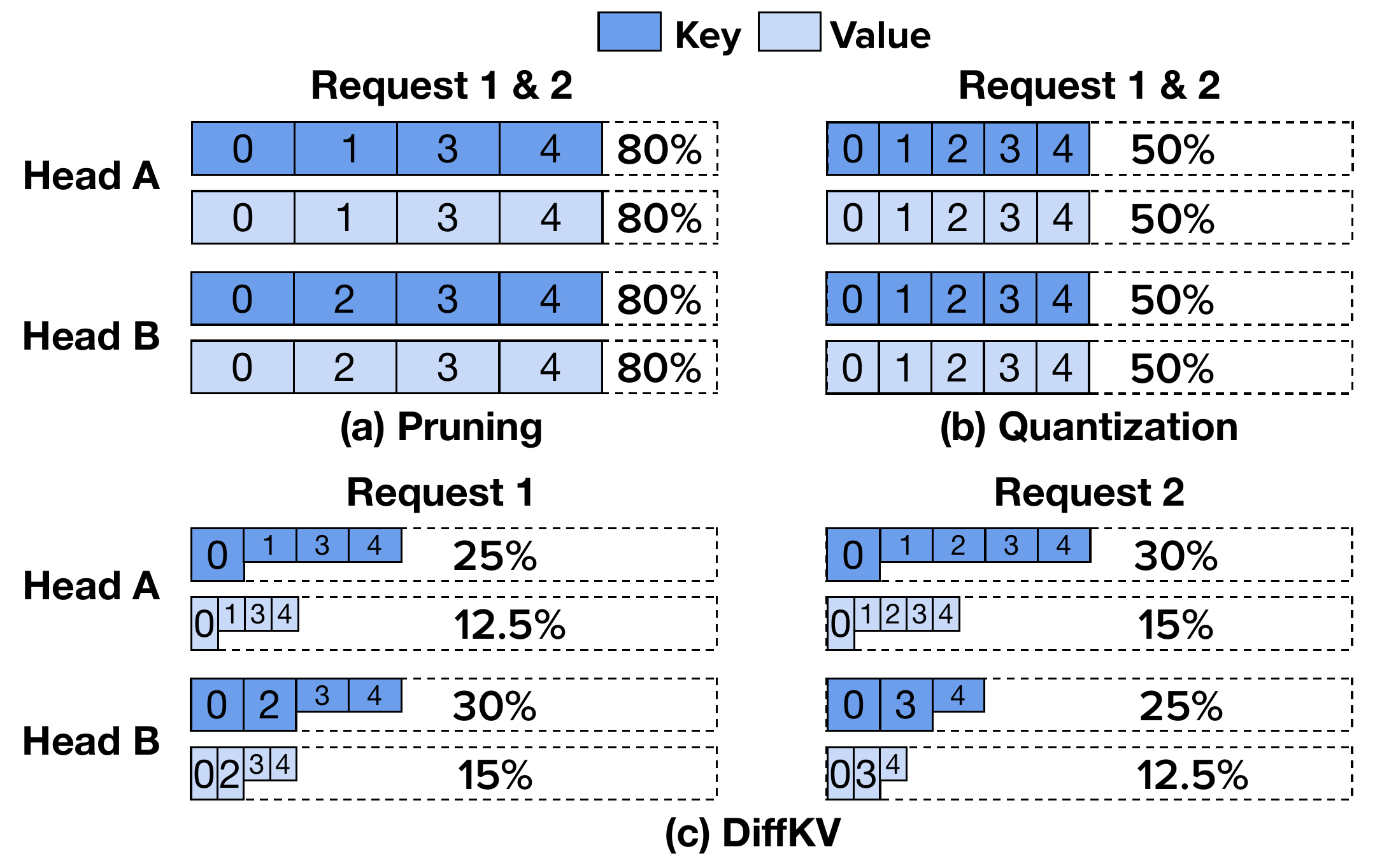}
\vspace{-0.1in}
\caption{KV cache memory allocation patterns for (a) pruning, (b) uniform quantization, and (c) \system across two attention heads in two 5-token requests. Boxes represent retained tokens (annotated with token IDs), with box size proportional to memory usage. Percentages indicate total memory usage relative to an uncompressed KV cache.}
\label{fig:mem_usage_patterns}
\vspace{-0.2in}
\end{figure}



Figure~\ref{fig:mem_usage_patterns} compares the KV cache usage of pruning, quantization, and \system across two heads in two 5-token requests. In both requests, pruning uniformly eliminates one token per head (reducing cache usage to 80\%), failing to adapt to heads with naturally sparser attention patterns. Quantization, meanwhile, applies INT8 precision uniformly to all FP16 values, halving memory usage but missing opportunities for finger-grained savings by treating keys and values differently based on their relative importance. Moreover, both approaches treat two requests identically, overlooking their distinct sparsity patterns.

In contrast, \system exhibits more adaptive memory usage. In Request 1, for Head A, \system prunes one token and further distinguishes precision across the rest: Token 0, deemed highly important, is stored at high precision (key in INT8, value in INT4), while Tokens 1, 3, and 4 are stored at lower precision (keys in INT4, values in INT2). This reduces usage to 25\% for keys and 12.5\% for values.
For Head B, \system instead identifies two tokens as highly important, resulting in 30\% and 15\% KV cache usage for keys and values, respectively. Averaged across both heads, \system uses 20.6\% of memory for Request 1, outperforming pruning and quantization.
Additionally, \system also adapts flexibly across requests. In Request 2, Head A stores one token at high precision and the remaining four at lower precision without pruning, whereas Head B prunes two tokens and stores two others at high precision. These examples highlight \system's ability to leverage both per-head dynamic sparsity and request-level variation, enabling more aggressive and flexible compression than prior methods.

Although \system achieves a high KV cache compression ratio, it introduces significant memory usage irregularity. As depicted in Figure~\ref{fig:mem_usage_patterns}, pruning and quantization maintain uniform KV cache utilization across attention heads. In contrast, \system exhibits substantial variability, not only across different heads but also between the key and value within individual heads.
To mitigate fragmentation, the memory manager must precisely track and allocate memory per head and per request. However, this introduces a scalability challenge: even a modestly sized LLM like \llama3-8B~\cite{dubey2024llama} comprises hundreds of attention heads, and with hundreds of concurrent requests, the memory manager must handle tens of thousands of heterogeneous memory regions during each inference step.
Considering that model execution time per step is on the order of tens of milliseconds, improper memory management could render the overhead of dynamic allocation prohibitive, negating the benefits of KV cache compression.

To overcome the scalability challenge, \system introduces an on-GPU memory manager that efficiently handles the irregular memory usage patterns via \textit{parallel KV compaction}, which optimizes memory allocation and recycling by packing fragmented free memory lists into contiguous regions directly on the GPU in parallel.
Parallel KV compaction is enabled by three essential on-GPU data structures:

\myparagraph{1. Unified Pages} 
GPU memory is partitioned into fixed-size pages, each storing tokens at a specific precision. 
For example, a page may hold tokens with keys in INT8 and values in INT4. Each page is dynamically configured and parsed according to the precision requirements of the head and request.

\myparagraph{2. Circular Free Page List}
All page IDs are managed in a centralized, GPU-resident circular list. Both free and used pages are kept in contiguous regions, enabling efficient page allocation and recycling via parallel prefix sum~\cite{harris2007parallel}.

\myparagraph{3. Bidirectional Page Table}
Instead of maintaining separate page tables for high- and low-precision pages, \system consolidates both into a single bi-directional page table to minimize metadata overhead. High-precision page IDs grow from the left side of the table, and low-precision page IDs grow from the right side. 
Additional precision levels can also be efficiently accommodated by employing multiple bi-directional page tables.


We have implemented \system on vLLM~\cite{kwon2023efficient} and evaluate it across multiple LLMs, including the emerging thinking models QwQ-32B~\cite{qwq32b}, R1-Distill-Qwen-14B and R1-Distill-Llama-8B~\cite{guo2025deepseek}. 
Our experiment results demonstrate that Diff-\\KV offers a superior cost-accuracy tradeoff. Specifically, Diff-\\KV compresses the KV cache by $2.7\times$ to $5.7\times$ with near-lossless accuracy, resulting in throughput improvement of $1.9\times$ to $5.4\times$.
Notably, \system is, to the best of our knowledge, the first KV cache compression framework evaluated on thinking models and complex reasoning tasks that require advanced chain-of-thought (CoT) capabilities, while achieving FP16-comparable generation quality.

\vspace{-0.05in}

\section{Background and Related Work}
\label{sec:background}


In this section, we describe Transformer-based large language models and review existing techniques for KV cache management.

\subsection{Large Language Models}

Large language models (LLMs) are predominantly built upon the Transformer architecture~\cite{vaswani2017attention}.
At the core of Transformer is the attention mechanism, which allows each token in a sequence to weigh the significance of other tokens when constructing its contextualized representation.
During autoregressive inference, the attention mechanism operates in a causal manner, attending only to preceding tokens.
Mathematically, the standard attention computation is defined as:
\begin{equation}
\begin{aligned}
    \operatorname{Attention}(\mathbf{Q}, \mathbf{K}, \mathbf{V})_i &= \sum_{j=1}^{i} \operatorname{softmax}\left( \frac{\mathbf{Q} \mathbf{K}^\top}{\sqrt{d}} \right)_{ij} \mathbf{v}_j \\
    &= \sum_{j=1}^{i} \frac{\exp\left( \frac{\mathbf{q}_i \cdot \mathbf{k}_j}{\sqrt{d}} \right)}{\sum_{n=1}^{i} \exp\left( \frac{\mathbf{q}_i \cdot \mathbf{k}_n}{\sqrt{d}} \right)} \mathbf{v}_j
\end{aligned}
\label{eq:attention}
\end{equation}
$\mathbf{Q}$, $\mathbf{K}$ and $\mathbf{V}$ are matrices of size $l \times d$, representing the queries, keys and values respectively, 
where $l$ denotes the number of tokens processed so far in the sequence and $d$ represents the feature dimensionality.
Vectors $\mathbf{q}_i$, $\mathbf{k}_i$, and $\mathbf{v}_i$ correspond to the query, key, and value of the $i^\text{th}$ token. 





To capture a broader range of interactions between tokens, Transformer models employ multi-head attention (MHA), where each head independently computes attention using distinct projections of queries, keys, and values.
Grouped-query attention (GQA) \cite{ainslie2023gqa, shazeer2019fast} improves the efficiency of MHA by allowing a group of query heads to share the same projection of keys and values, referred to as a KV head.


LLM execution involves two phases: the \textit{prompt phase}, where the model computes latent representations for all tokens in the prompt and generates the first new token, and the \textit{generation phase}, where subsequent tokens are generated iteratively.
To avoid redundant computations across generation steps, \textit{KV cache} is introduced to store the keys and values of all previous tokens.
However, the size of the KV cache grows linearly with both the sequence length and batch size, quickly becoming a bottleneck for inference throughput~\cite{yu2022orca, kwon2023efficient}.
Therefore, efficient KV cache management is critical for alleviating memory bottlenecks and improving LLM serving efficiency.





\subsection{KV Cache Optimization}

\myparagraph{PagedAttention}
Static KV cache management systems \cite{yu2022orca, FasterTransformer} reserve memory for the maximum possible sequence length, leading to considerable memory waste when actual sequences are shorter.
To mitigate this inefficiency, vLLM~\cite{kwon2023efficient} introduces \textit{PagedAttention}, which partitions the KV cache into pages, each containing a fixed number of tokens, and allocates the pages on-demand as the sequence length grows.
By managing the KV cache at the granularity of pages, vLLM reduces memory waste and enables larger batch sizes to improve serving efficiency.

\myparagraph{KV Cache Quantization}
Quantization~\cite{nagel2021white, dettmers2022gpt3, guo2023olive, hooper2024kvquant} reduces the KV cache size by approximating high-precision floating points with discrete low-bit values.
For a tensor $\mathbf{X}$, we first compute the scale $s$ and zero point $z$ based on $\mathbf{X}_{min}$ and $\mathbf{X}_{max}$, then apply quantization element-wise as: $\mathbf{Q} = \operatorname{round}\left( \frac{\mathbf{X} - z}{s} \right)$.
During inference, the original tensor is approximately reconstructed via dequantization: $ \hat{\mathbf{X}} = s \cdot \mathbf{Q} + z $.
Unlike the quantized tensor $\mathbf{Q}$, the metadata $s$ and $z$ are kept in higher precision (e.g., FP16) to ensure more accurate dequantization.
State-of-the-art KV cache quantization methods such as Atom \cite{zhao2024atom} and Qserve \cite{lin2024qserve} apply this process to each key and value vector independently.
However, these uniform quantization methods, which apply a fixed bit-width across keys and values of all tokens, may not be optimal. They overlook the varying importance of tokens as well as the different roles of keys and values in the attention calculation.



\myparagraph{KV Cache Pruning} 
Several recent works~\cite{zhang2024h2o, xiao2023efficient, li2024snapkv, zhang2024pyramidkv, liu2024scissorhands, ge2023model}
have explored KV cache pruning to alleviate the memory bottleneck in LLM inference, which can be considered as an extreme case of quantization.
These methods use attention scores to assess token importance and evict less important tokens, thereby reducing the KV cache size.
The effectiveness of these methods, however, is limited by their static and inflexible allocation of memory resources.
Specifically, H2O~\cite{zhang2024h2o} and SnapKV~\cite{li2024snapkv} allocate the same memory budget uniformly across all attention heads and layers.
PyramidKV~\cite{zhang2024pyramidkv} allocates more memory in lower layers and less in higher ones, based on the empirical observation that the number of important tokens shrinks as the model depth increases, but still relies on static heuristics and cannot dynamically adjust cache allocations to suit various workloads.
Complementary to KV cache pruning, 
another line of work~\cite{sheng2023flexgen, lee2024infinigen} addresses GPU memory constraints by offloading the entire KV cache to CPU memory and selectively loading important tokens to the GPU for attention computations.
However, these methods do not inherently reduce the KV cache size and incur additional latency due to data transfers between the CPU and GPU.




\section{The Case for Differentiated KV}
\label{sec:motivation}

We motivate the design of \system by presenting the case for three levels of differentiation within the KV cache: (i) the differentiated impacts of keys and values (Section~\ref{sec:impact_kv}), (ii) differentiated token importance (Section~\ref{sec:diff_token_important}), and (iii) per-head dynamic sparsity (Section~\ref{sec:dynamic_sparsity}). Building on these insights, Section~\ref{sec:diffkv_insights} describes how DiffKV combines them together and highlights its contributions over prior systems.

\vspace{-0.1in}

\subsection{Differentiated Impacts of Keys and Values}
\label{sec:impact_kv}

While the significance of different tokens can be directly inferred from attention scores, the roles of the key and value vectors within a token are less obvious. 
However, examining Equation~\ref{eq:attention} reveals a divergence in their impacts. Each token's contribution to the attention output depends on two factors: the attention score from softmax and the value vector.
Key vectors, as part of the shared softmax denominator, influence the attention scores of all tokens, whereas value vectors only affect their respective token's contribution to the output. 

Beyond the broader impact of key vectors, it is crucial to establish the \textit{relative significance} of attention scores, determined by key vectors, over value vectors. Without this, the significance of key vectors could be overshadowed by that of value vectors.
To this end, we reformulate the attention mechanism as a weighted sum of unit vectors, as presented in Equation~\ref{eq:attention_rewrite}.
This formulation decomposes each input token's contribution into two components: the unit vector $\frac{v_j}{|v_j|}$, 
obtained by dividing the token's value vector by its L2 norm to capture only its direction in feature space,
and a coefficient, defined as the product of the attention score and the norm of the value vector, which determines the relative significance of the token's direction. 
This allows us to assess the relative importance of attention scores and value vectors by examining their contributions to these coefficients.
\begin{equation}
\begin{aligned}
    \operatorname{Attn}(\mathbf{Q}, \mathbf{K}, \mathbf{V})_i &= \sum_{j=1}^{i} 
    \underbrace{\operatorname{softmax}\left( \frac{\mathbf{Q} \mathbf{K}^\top}{\sqrt{d}} \right)_{ij} |\mathbf{v}_j|
    }_{\text{Coefficient}} 
    \underbrace{ \frac{\mathbf{v}_j}{|\mathbf{v}_j|}}_{\text{Unit vector}}
\end{aligned}
\label{eq:attention_rewrite}
\end{equation}

\begin{figure}[t]
\centering
\includegraphics[width=.95\linewidth]{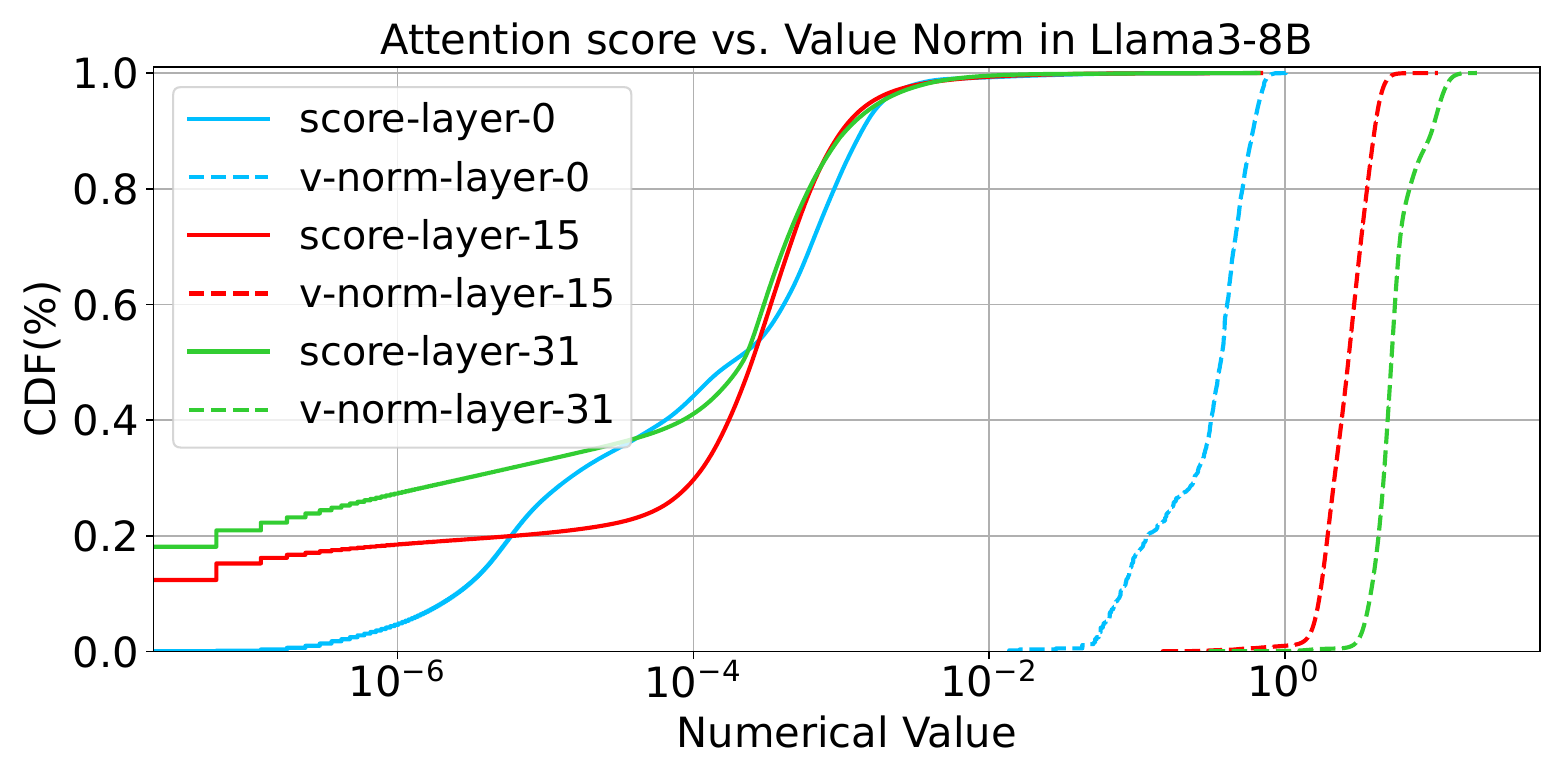}
\vspace{-0.1in}
\caption{Distribution of attention score and value vector norm in \llama3-8B.}
\vspace{-0.4em}
\label{fig:attn_score_vs_v_norm}
\end{figure}

Figure~\ref{fig:attn_score_vs_v_norm} presents the distributions of the average attention score per token and the norms of value vectors across three representative layers of the \llama3-8B model~\cite{dubey2024llama}, evaluated on the first one thousand articles from the Wikitext dataset~\cite{merity2016pointer}. 
Notably, the attention scores span seven orders of magnitude, vastly exceeding the range of value vector norms, which only cover at most two orders of magnitude. Similar patterns are also observed in the larger \llama3-70B model. 
This pronounced disparity highlights the pivotal role of attention scores in determining each token's contribution to the attention output. We therefore conclude that key vectors exert a broader and more impactful influence than value vectors, motivating further exploration of processing key and value vectors at distinct precision levels.
We validate this intuition in Section~\ref{sec:eval-accuracy}, showing that K8V4 and K4V2, which quantize keys to 8/4 bits and values to 4/2 bits respectively, outperform their mirror configurations K4V8 and K2V4.


\subsection{Differentiated Token Importance}
\label{sec:diff_token_important}


Tokens contribute to attention outputs with varying degrees of importance, as reflected by their attention scores.
By exploiting these differences, we can apply finer-grained compression strategies that go beyond uniformly quantizing all tokens \cite{zhao2024atom, lin2024qserve} or exclusively pruning the least important ones \cite{zhang2024h2o, zhang2024pyramidkv}.

\begin{figure}[]
\centering
\includegraphics[width=1.0\linewidth]{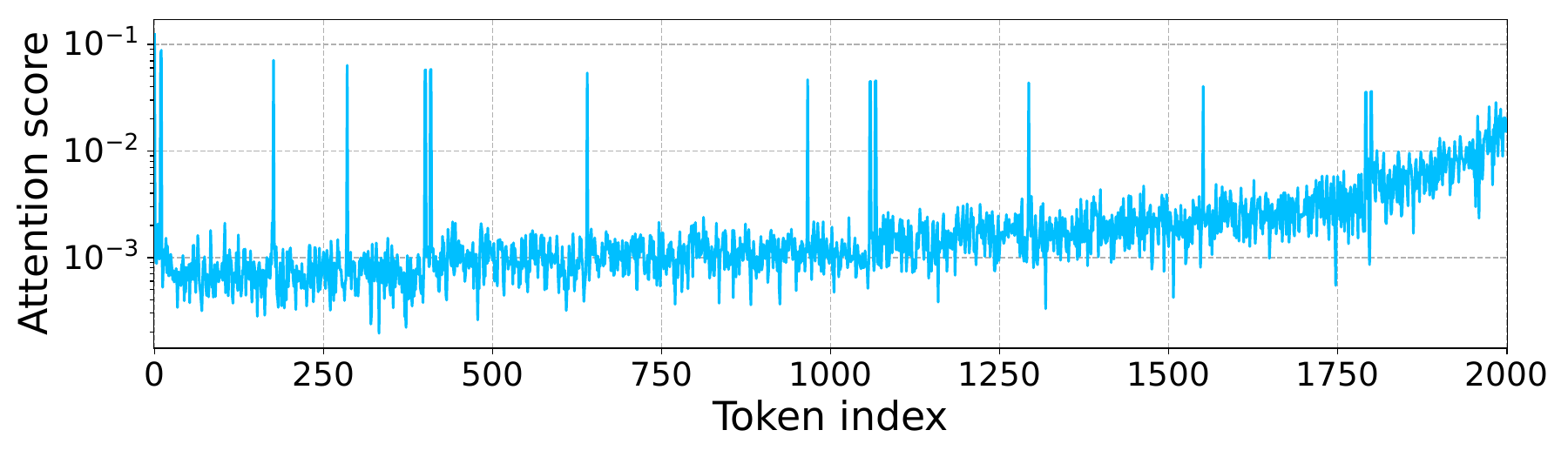}
\vspace{-2em}
\caption{Per-token attention scores in the $8^{th}$ layer of \llama3-8B on a sequence randomly sampled from Wikitext.}
\label{fig:score_dist_seq}
\vspace{-1em}
\end{figure}

Figure~\ref{fig:score_dist_seq} shows the per-token attention scores in the $8^{th}$ layer, selected as a representative layer of \llama3-8B~\cite{dubey2024llama}, on a sequence randomly sampled from the Wikitext dataset~\cite{merity2016pointer}. Similar sparsity patterns are observed across the other layers as well.
This non-uniform distribution of token importance motivates a \emph{hierarchical} compression strategy that allocates memory based on token significance: high-precision storage (e.g., K8V4) for the most important tokens, lower precision (e.g., K4V2) for moderately important tokens, and pruning for the least important.
This hierarchical approach enables a smoother trade-off between memory savings and generation quality than uniform quantization methods (e.g., Atom~\cite{zhao2024atom}, QServe~\cite{lin2024qserve}) and pruning-only approaches (e.g., H2O~\cite{zhang2024h2o}, SnapKV~\cite{li2024snapkv}).

\subsection{Per-head Dynamic Sparsity Patterns}
\label{sec:dynamic_sparsity}


LLMs exhibit dynamic sparsity patterns that vary across both attention heads and requests: the number of critical tokens can differ not only between heads but also for the same head under different requests.
To characterize the sparsity of attention, we analyze the minimum number of critical tokens required to retain the majority of information, specifically by preserving a target percentage (e.g., 95\%) of the total attention score.

\begin{figure}[t]
\centering
\includegraphics[width=.95\linewidth]{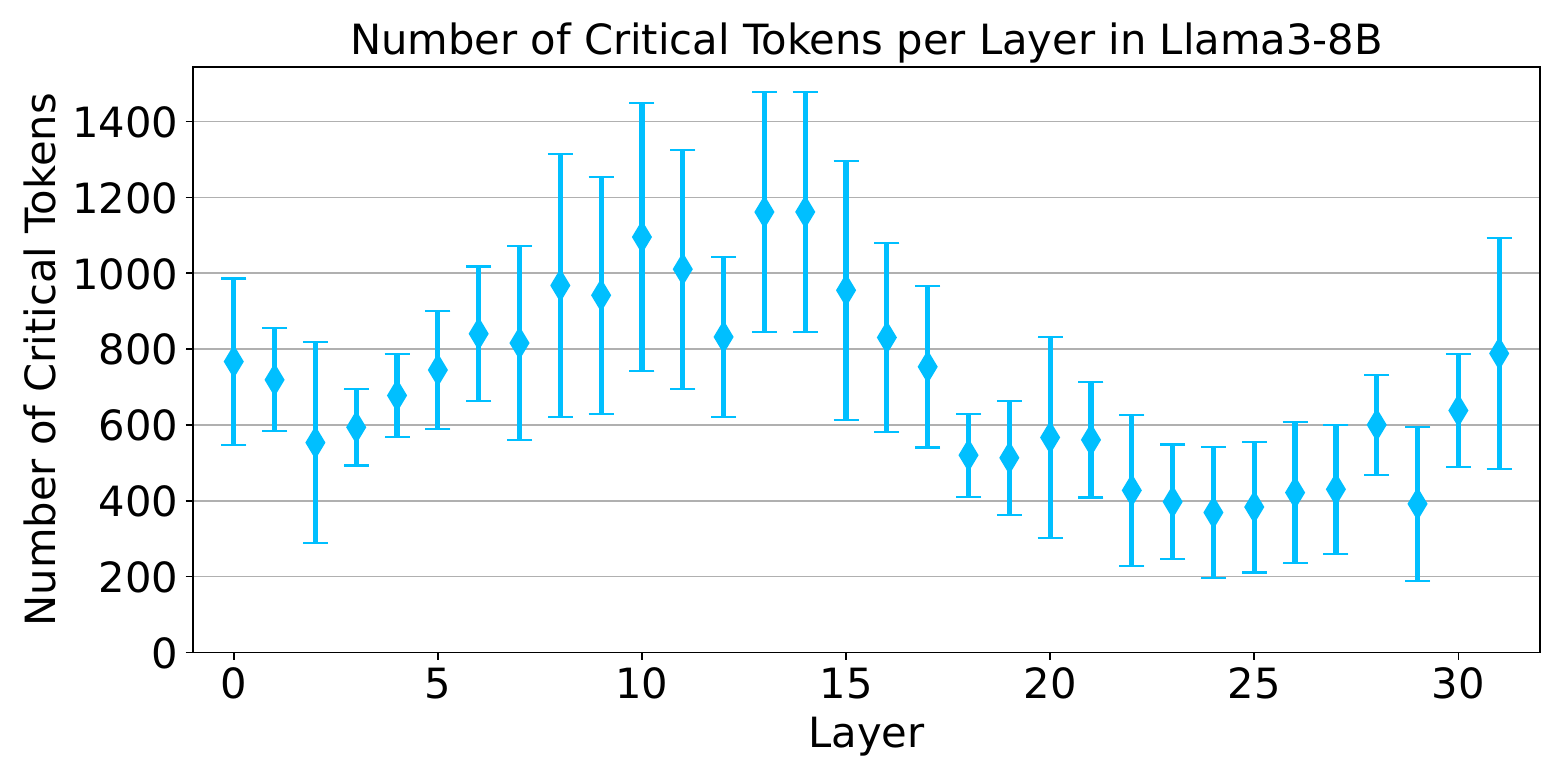}
\vspace{-0.16in}
\caption{Number of critical tokens per layer in \llama3-8B to preserve 95\% of the total attention score.}
\label{fig:sparsity_per_layer}
\vspace{-0.2in}
\end{figure}

We begin by investigating dynamic sparsity patterns across layers, and evaluate the \llama3-8B model
with the first thousand articles from the Wikitext dataset. Figure~\ref{fig:sparsity_per_layer} illustrates the average number of critical tokens per layer, aggregated across all KV heads, required to retain 95\% of the total attention score. Vertical bars represent the standard deviation, capturing variability across individual requests. Notably, the degree of sparsity varies considerably across layers, as indicated by differences in the number of critical tokens.

\begin{figure}[t]
\centering
\includegraphics[width=.95\linewidth]{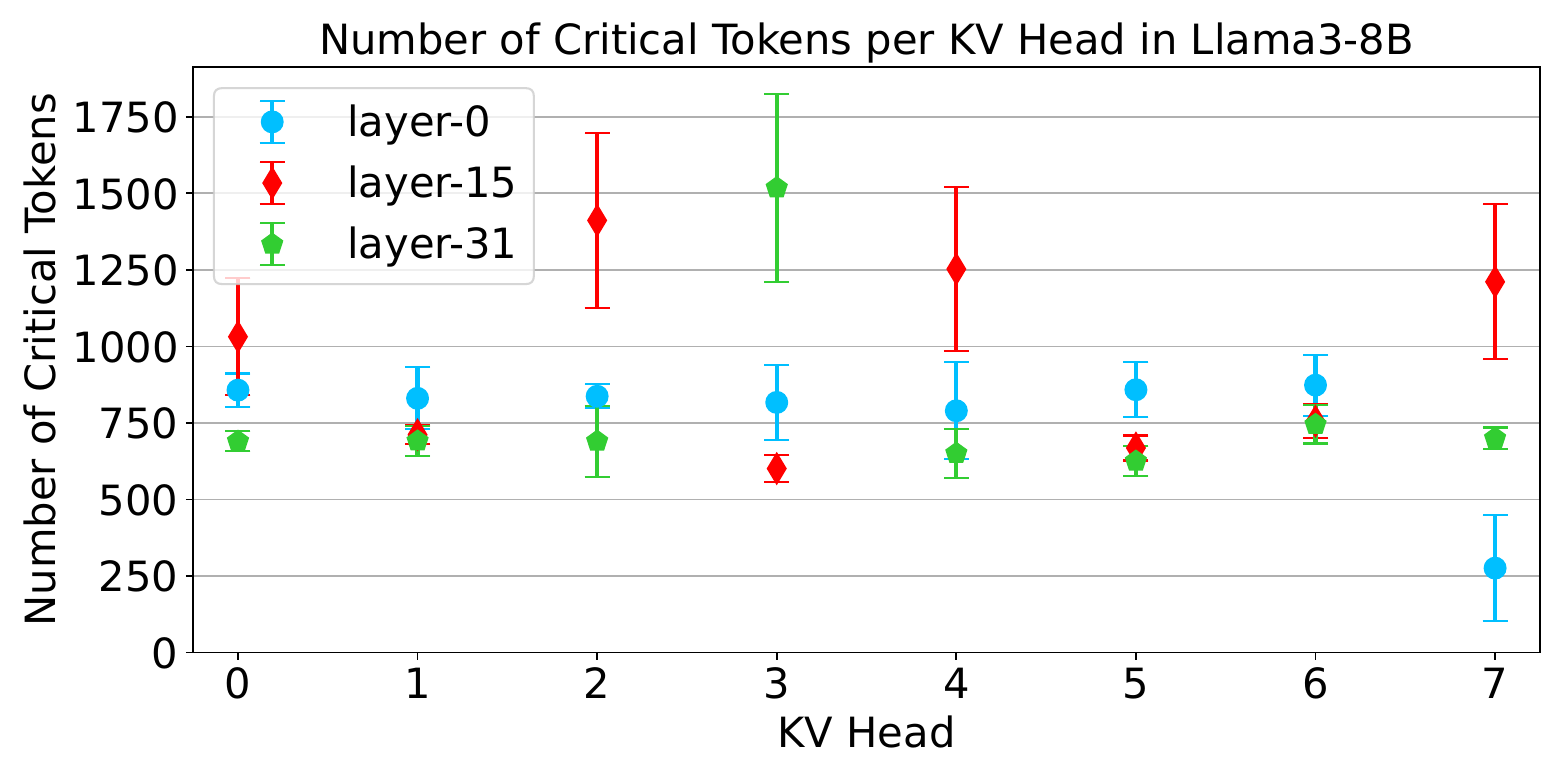}
\vspace{-0.16in}
\caption{Number of critical tokens per KV head in \llama3-8B to preserve 95\% of the total attention score.}
\label{fig:sparsity_per_head}
\vspace{-0.2in}
\end{figure}

Next, we delve into the dynamic sparsity within individual layers. Figure~\ref{fig:sparsity_per_head} presents the average number of critical tokens per KV head for three representative layers, with horizontal bars indicating the standard deviation across requests. The sparsity pattern remains highly dynamic: within each layer, the number of critical tokens varies significantly across KV heads. Furthermore, within individual KV heads, the number of critical tokens can vary substantially across requests, as shown by the high standard deviation in some KV heads.

\vspace{-0.1in}

\subsection{Main Insights and Implications}
\label{sec:diffkv_insights}



Our findings reveal three critical levels of differentiation within the KV cache:

\vspace{-0.08in}
\begin{itemize}
\item Keys exert a broader impact on attention computation than values, motivating differentiated precisions for keys and values.
\vspace{-0.08in}
\item Tokens vary in importance as reflected by their attention scores, motivating a fine-grained, hierarchical compression strategy.
\vspace{-0.08in}
\item Attention sparsity patterns vary across requests and attention heads, requiring dynamically managing memory resources on a per-request and per-head basis.
\end{itemize}
\vspace{-0.08in}

Collectively, these highly dynamic sparsity patterns, across layers, attention heads, and individual requests, underscore the necessity of adaptive memory management on a per-head and per-request basis. They motivate the design of \system, which integrates differentiated precisions for keys and values, hierarchical compression based on importance, and per-request and per-head memory management.

\myparagraph{Contributions of \system}
\system expands the design space for KV cache compression by \textit{jointly exploiting the three levels of differentiation in KV cache}, enabling higher compression ratios with minimal quality degradation.
Compared to prior work that explores importance-based KV cache pruning~\cite{zhang2024h2o,li2024snapkv,adnan2024keyformer,tang2024quest} and mixed-precision quantization~\cite{zhao2024atom,dong2024qaq,he2024zipcache}, \system introduces the following key innovations:
\vspace{-0.08in}
\begin{itemize}
\item \textit{Request-aware differentiated KV quantization.} 
\system adaptively adjusts the mix of high- and low-precision tokens (e.g., K8V4, K4V2) on a per-request basis (Section~\ref{sec:unified_kv_compression}), whereas prior work~\cite{zhao2024atom,dong2024qaq,adnan2024keyformer} typically applies a single static configuration across all requests.
\vspace{-0.08in}
\item \textit{Sequence length-adaptive importance estimation.} 
Diff-KV modulates the proportion of important tokens based on sequence length, retaining more tokens in shorter sequences to preserve quality, while enabling more aggressive compression for longer ones (Section~\ref{sec:unified_kv_compression}). Prior approaches~\cite{zhang2024h2o,li2024snapkv,adnan2024keyformer} typically use a fixed token retention ratio, regardless of sequence length.
\vspace{-0.08in}
\item \textit{Per-head dynamic sparsity exploitation}.
\system dynamically adjusts memory allocation per head, based on observed sparsity patterns (Section~\ref{sec:dynamic_sparsity} \& \ref{sec:unified_kv_compression}), whereas existing methods allocate memory uniformly across attention heads, missing optimization opportunities.
\vspace{-0.08in}
\item \textit{Scalable on-GPU memory management}.
\system introduces an efficient on-GPU memory manager that handles irregular per-head memory allocation patterns, effectively translating memory savings into performance gains  (Section~\ref{sec:system_design}).
\end{itemize}
\vspace{-0.08in}

\vspace{-0.05in}

\section{KV Compression Policy}
\label{sec:unified_kv_compression}

\vspace{-0.05in}




We present the KV compression policy of \system, designed to address the three levels of differentiation within the KV cache.
First, to reflect the greater impact of keys on attention computation compared to values, we propose quantizing keys at a higher precision than values.
For example, keys and values can be quantized to 8 and 4 bits (K8V4), or 4 and 2 bits (K4V2).
Second, to account for the varying importance of tokens, we introduce a hierarchical compression strategy that classifies tokens into three significance levels: the most important tokens are quantized at high precision (e.g., K8V4), moderately important tokens at lower precision (e.g., K4V2), and the least important tokens are pruned entirely.
This strategy is both request-aware and sequence-length adaptive. It dynamically adjusts the mix of high- and low-precision tokens per request to reflect their varying sparsity patterns. Moreover, it retains a larger fraction of tokens at high precision for short sequences to preserve quality, while applying low precision quantization and pruning more aggressively for long sequences to maximize memory savings.

Finally, to address the dynamic attention sparsity patterns across requests and attention heads, we propose an adaptive memory management approach.
Rather than imposing a fixed memory budget, our approach allows each attention head to determine its memory requirements dynamically based on its specific sparsity pattern.


Next, we describe the KV compression policy in greater detail, addressing the prompt phase and the generation phase separately.
In both phases, compression is applied per-request and per-head, ensuring that memory usage is tailored to the specific sparsity pattern of each request and attention head.

\myparagraph{Prompt Phase}
In the prompt phase, key and value vectors for all tokens in the prompt are computed. The compression policy then determines the appropriate precision for storing each token based on its significance. The significance of the $i^{th}$ token is calculated by averaging the $N - i$ attention scores it receives from subsequent tokens, where $N$ denotes the prompt sequence length. 
In the case of GQA and MHA, scores from all attention heads associated with the KV head are aggregated using the maximum operation.
To mitigate premature compression, the most recent $W$ tokens are always quantized at high precision, where $W$ is typically set to 64. 
For the remaining tokens, the policy determines the precision level of the $i^{th}$ token in a sequence-length adaptive manner, 
by comparing its significance score to the theoretical average $\frac{1}{i}$, based on the intuition that tokens with scores below average are less critical.
Specifically, the $i^{th}$ token is quantized at high precision if its significance score exceeds $\frac{\alpha_h}{i}$, at low precision if its score lies within the interval $[\frac{\alpha_l}{i}, \frac{\alpha_h}{i}]$, and is pruned otherwise. The parameters $\alpha_l$ and $\alpha_h$, which define the thresholds for high- and low-precision quantization, are determined offline through profiling on a calibration dataset.
As a result of this hierarchical compression, the KV cache is conceptually divided into two parts: a high-precision section $KV_h$ and a low-precision section $KV_l$. 

\myparagraph{Generation Phase}
In the generation phase, a single token is compressed at each step, aligning with the autoregressive nature of the generation process. 
The most recent token is added to the recent window to prevent premature compression, while the earliest token $t_c$ in the window becomes a candidate for more aggressive compression. 
The compression procedure can be divided into two parts. First, token $t_c$ is quantized at either a high or low precision and added to the corresponding section of the KV cache, or it may be completely pruned. Next, if $t_c$ is quantized, the least significant token $t_v$ in the corresponding precision section of the KV cache is considered for further downgrading: it may be re-quantized to lower precision or pruned, depending on its significance.
Essentially, this policy establishes a smooth downgrading path for the less important token: rather than being pruned directly, it is first re-quantized to low precision, with pruning occurring only if it remains insignificant.

The detailed procedure is outlined in Algorithm~\ref{algo:compression_generation}, following the same intuition and parameters as the prompt phase. Given the sequence length $N$, $t_c$ is quantized at high precision and added to $KV_h$, the high precision KV cache (line~\ref{lst:line:q_high}), if its significance exceeds $\frac{\alpha_h}{N}$.
Subsequently, the least significant token in $KV_h$ is identified as the victim token $t_v$ for more aggressive compression (line~\ref{lst:line:v_high}). If $t_v$'s significance exceeds $\frac{\alpha_h}{N}$, it remains in $KV_h$; if it falls within $[\frac{\alpha_l}{N}, \frac{\alpha_h}{N}]$, $t_v$ is re-quantized to low precision and moved to $KV_l$, the low precision KV cache (line~\ref{lst:line:requant_high}); otherwise, $t_v$ is pruned.
Similarly, if $t_c$'s significance lies within $[\frac{\alpha_l}{N}, \frac{\alpha_h}{N}]$, it is quantized to low precision and added to $KV_l$ (line~\ref{lst:line:q_low}). The least important token in $KV_l$ is designated as the victim $t_v$ (line~\ref{lst:line:v_low}), and is further pruned if its significance falls below $\frac{\alpha_l}{N}$ (line~\ref{lst:line:prune_low}).


\myparagraph{Discussion}
The proposed KV compression policy is highly extensible, allowing for the potential use of more than two quantization precision levels.
We adopt two quantization precision levels primarily to minimize metadata overhead and improve system efficiency, as will be detailed in the following section.
In Section~\ref{sec:eval-accuracy}, we empirically show that the two-level scheme (K8V4-K4V2) achieves near-lossless generation quality across multiple models and benchmarks.
We use a shared set of thresholds for all attention heads, as our empirical studies demonstrate that this approach is sufficient to capture the varying sparsity patterns across different heads.
Notably, \system can efficiently support more flexible compression policies, such as tuning thresholds for each head individually, which could potentially further optimize the balance between model accuracy and memory efficiency.
We leave the exploration of such flexible policies for future work.

\begin{algorithm}[!h]
\SetAlgoLined
\begin{algorithmic}[1]
\STATE \textbf{Input:} Parameters $\alpha_h$, $\alpha_l$; High \& low precision $P_h$ \& $P_l$
\STATE \textbf{Input:} Candidate token $t_c$; Sequence length $N$
\STATE \textbf{Input:} High \& low precision KV cache $KV_h$ \& $KV_l$
\STATE \textbf{Function:} Significance $Score$; Quantization $Quant$;
 \IF {Score($t_c$) $\geq$ $\frac{\alpha_h}{N}$}
    \STATE $KV_h$.add(Quant($t_c$, $P_h$))  \label{lst:line:q_high}
    \STATE $t_v$ = $argmin_{t \in KV_h}$(Score($t$)) \label{lst:line:v_high}
    \IF {$\frac{\alpha_l}{N} \leq$ Score($t_v$) < $\frac{\alpha_h}{N}$} 
        \STATE $KV_h$.remove($t_v$), ~$KV_l$.add(Quant($t_v$, $P_l$))  \label{lst:line:requant_high}
    \ELSIF {Score($t_v$) < $\frac{\alpha_l}{N}$} 
        \STATE $KV_h$.remove($t_v$)
    \ENDIF
 \ELSIF{Score($t_c$) $\geq$ $\frac{\alpha_l}{N}$}
    \STATE $KV_l$.add(Quant($t_c$, $P_l$)) \label{lst:line:q_low}
    \STATE $t_v$ = $argmin_{t \in KV_l}$(Score($t$)) \label{lst:line:v_low}
    \IF{Score ($t_v$) < $\frac{\alpha_l}{N}$}
        \STATE $KV_l$.remove($t_v$) \label{lst:line:prune_low}
    \ENDIF
 \ENDIF
 \end{algorithmic}
 \caption{KV compression policy (generation)}
 \label{algo:compression_generation}
 \vspace{-0.0305in}
\end{algorithm}
\vspace{-0.1in}




\vspace{-0.09in}

\section{Memory Management}

\vspace{-0.05in}


\label{sec:system_design}

In this section, we describe \system's memory management mechanism which efficiently supports the three levels of differentiation in the KV cache.
We first highlight the challenges posed by these differentiations in memory management, and then present
parallel KV compaction, a novel memory management technique that effectively tackles the challenges.

\vspace{-0.05in}

\subsection{Challenges}

\myparagraph{Flexible Paging}
In PagedAttention~\cite{kwon2023efficient}, all tokens are stored at the same precision, allowing for a fixed page format. However, the differentiation of key and value vectors, along with varying token importance, introduces multiple precision levels both within and across tokens, making a fixed page format insufficient.
Specifically, a fixed page format requires conservatively allocating high-precision slots for all tokens regardless of their actual precisions, leading to considerable memory wastage.
Suppose memory is allocated to accommodate the highest precision K8V4, a token with K4V2 precision would waste 50\% of the memory.
Worse, such a fixed page format results in misaligned memory accesses, which hinder memory bandwidth utilization and reduce overall computational efficiency.

\myparagraph{KV Compaction Scalability}
At each LLM inference step, the memory management complexity is $O(\#requests)$ for PagedAttention, as memory is partitioned uniformly among attention heads.
In contrast, \system must handle varying numbers of high- and low-precision tokens across different heads, mapping these heterogeneous memory requirements to physical memory. We term this process \textit{KV compaction}, whose complexity is $O(\#requests \times \#heads)$ per inference step. This poses significant scalability challenges for memory management.
Given that model execution time per step is on the order of tens of milliseconds, improper memory management
could render the overhead of dynamic allocation prohibitive,
negating the benefits of KV cache compression.
Moreover, managing both high-precision and low-precision tokens would require separate data structures for each precision, resulting in increased metadata overhead.



\vspace{-0.1in}

\subsection{Parallel KV Compaction}

A detailed analysis of the KV compaction process reveals opportunities for parallelization. KV compaction can be divided into two phases: \textit{planning} and \textit{coordination}.
In the planning phase, each attention head independently determines its memory allocation requirements. The subsequent coordination phase synchronizes these per-head requirements and maps them to the GPU's physical memory. 
The primary source of increased memory management complexity lies in the planning phase, which is perfectly parallelizable and well-suited to the GPU's parallel compute capabilities. Parallelizing the coordination phase is more challenging, as it requires synchronization across heads. However, we observe that this phase can be parallelized effectively via parallel prefix sum~\cite{nakano2012prefix-sum}, provided that the free memory region is contiguous.

Thus, we propose \textit{parallel KV compaction}, a novel management technique that efficiently performs per-head dynamic memory allocation and recycling in parallel directly on the GPU.
Parallel KV compaction is enabled by three GPU-resident data structures collectively: the $(1)$ \textit{unified pages} to enable flexible paging, $(2)$ the \textit{circular free page list} to efficiently parallelize memory management, and $(3)$ the \textit{bidirectional page table} to minimize metadata overhead for tracking tokens of differentiated precisions.



\myparagraph{Unified Pages}
\textit{Unified pages} abstract away the complexity of differentiated precisions both within individual tokens and across different tokens, thereby simplifying the implementation of parallel KV compaction.
Specifically, we partition the available GPU memory into evenly sized pages, each configured \textit{upon allocation} to store tokens at a given precision.
Each unified page is organized into six segments: quantized keys, quantization metadata for keys, quantized values, quantization metadata for values, token scores, and positions.
The quantization metadata includes scales and zero points for the key and value vectors.
The number of tokens stored per page is adjusted according to the quantization configuration, ensuring compact memory usage.
Furthermore, by consolidating keys, values, and their metadata into a single structure, unified pages enhance data locality and eliminate the need for scattered lookups, improving memory access efficiency during attention computation.





\myparagraph{Circular Free Page List}
The \textit{circular free page list} serves as the cornerstone of parallel KV compaction, facilitating the parallelization of memory allocation and recycling by maintaining both free and used pages in contiguous regions.

This centralized, GPU-resident data structure contains all Page IDs and tracks free pages through a pair of pointers: a start pointer for allocations and an end pointer for recycling. These pointers wrap around to the beginning of the list upon reaching the end, forming a circular structure. When a page is allocated, the start pointer advances to the next available Page ID; when a page is freed, the end pointer advances to add the released Page ID back into the list. 
Both the available and unavailable regions in the list remain contiguous, enabling the coordination phase in memory management to be parallelized via parallel prefix sum.

In parallel KV compaction, after each head determines the number of pages to be allocated or freed, a parallel prefix sum operation computes a unique offset for each head relative to the start or end pointer.
This step converts per-head memory demands into a globally consistent schedule, where each head is assigned a disjoint region of the free page list to read from (for allocation) or write to (for recycling), ensuring non-conflicting operations within the list.
For memory allocation, each head concurrently retrieves its new page IDs from its designated region in the list, with the start pointer incremented by the cumulative number of pages required across all heads.
Similarly, for memory recycling, each head writes freed page IDs to its designated region, with the end pointer incremented by the total number of pages released.





\myparagraph{Bidirectional Page Table}
Similar to PagedAttention, Diff-KV maintains a page table for each attention head, mapping each request to its corresponding list of page IDs. 
To avoid the doubled metadata overhead associated with maintaining separate page tables for high- and low-precision pages, \system introduces a unified, GPU-resident data structure called the \textit{bidirectional page table}. 
This structure efficiently supports the use of two quantization precision levels for KV cache, as proposed in Section~\ref{sec:unified_kv_compression}, minimizing the metadata overhead.
In each entry of the bidirectional page table, high-precision page IDs grow from the left side of the list, while low-precision page IDs grow from the right, dynamically adapting to the precision requirements of the workload. The length of each page table entry is determined by the maximum sequence length divided by the tokens per high-precision page, ensuring no overflow since low-precision pages always contain more tokens than high-precision ones.
This unified approach not only minimizes metadata overhead but also eliminates the need for separate lookups based on precision levels, thereby enhancing memory access efficiency during attention computations. The memory overhead of the bidirectional page table is minimal: for example, with a batch size of 128 on \llama3-8B, which has 32 layers and 8 KV heads per layer, the total size of all bidirectional page tables is only 32 MB. By contrast, the KV cache for a single request occupies 1 GB.


\vspace{-0.1in}

\subsection{KV Compaction Workflow}

We delve into the KV compaction workflow during the prompt and generation phases, illustrating how the three data structures interact to enable differentiated KV cache compression.
Notably, KV compaction is executed once per inference step for all requests in the batch and all layers in the LLM. This design ensures sufficient parallelism for efficient GPU execution, while amortizing the associated GPU kernel launch overheads.


\begin{figure}
\centering
\includegraphics[width=1\linewidth]{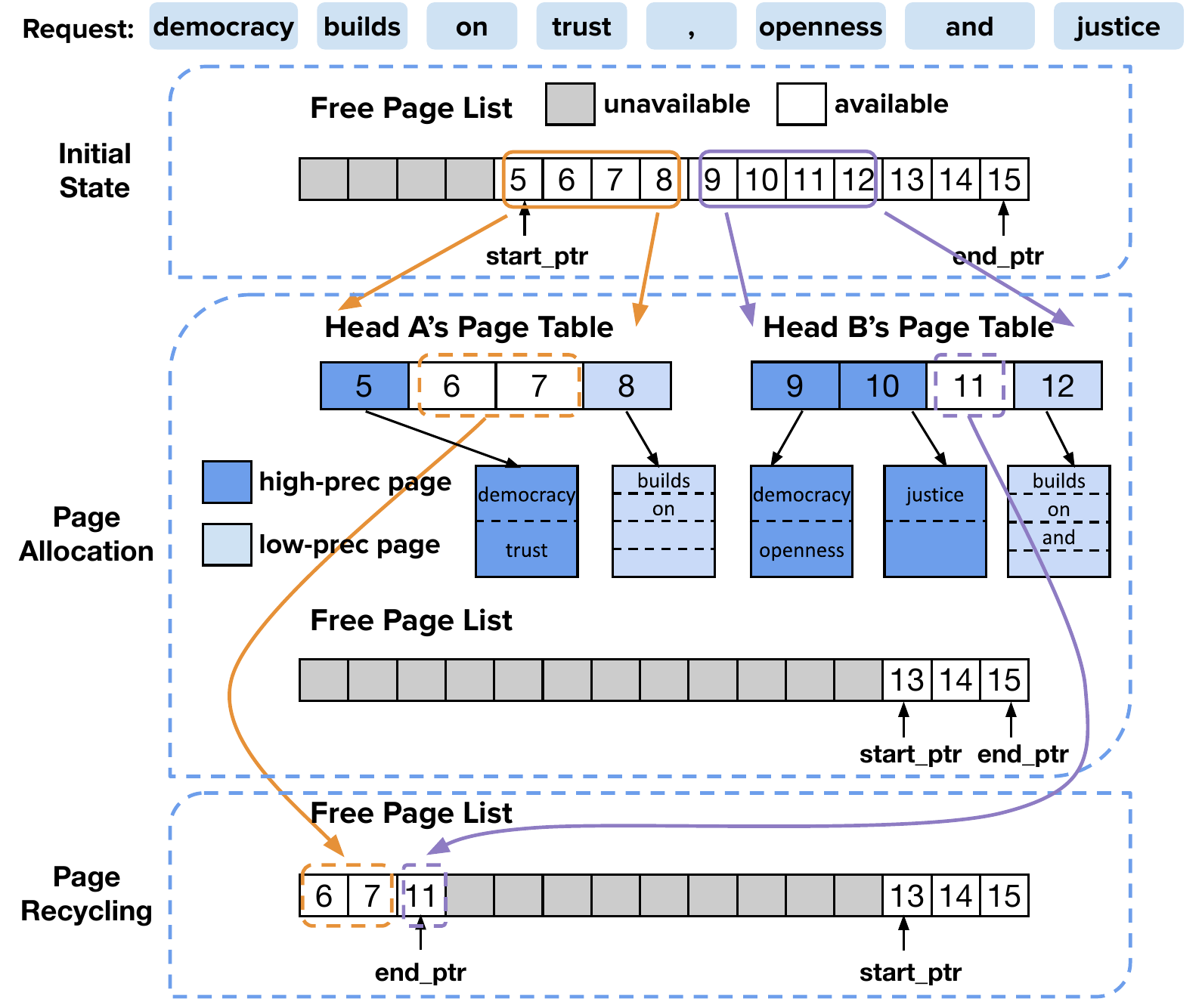}
\vspace{-1em}
\caption{Memory management flow in the prompt phase.}
\vspace{-0.21in}
\label{fig:memory_management_flow}
\end{figure}

\myparagraph{Prompt Phase}
Figure \ref{fig:memory_management_flow} illustrates the KV compaction workflow during the prompt phase for an example request with eight tokens.
In this example, one high-precision page stores two tokens, and one low-precision page stores four.
Since the exact numbers of high- and low-precision pages required by each head are unknown a priori, we conservatively allocate four unified pages per head, assuming all tokens are stored at high precision. Pages 5-8 and 9-12 are thus assigned to Head A and B, respectively, as shown in their page tables. The end pointer of circular free page list advances to Page 13, marking Pages 5-12 as allocated.

Next comes the \textit{planning phase}, during which each head independently applies the KV cache compression algorithm and calculates its specific memory requirements.
In this example, Head A uses one high-precision page, while Head B uses two; both heads also require one low-precision page.
We allocate high-precision pages left-to-right in the bidirectional page table and low-precision pages right-to-left: Head A uses Page 5 (high-precision) and Page 8 (low-precision), while Head B uses Pages 9–10 (high-precision) and Page 12 (low-precision).

Following the planning phase, the coordination phase finally reclaims unused pages (Pages 6-7 from Head A and Page 11 from Head B) via a parallel prefix-sum. These recycled pages are appended to the circular free page list; since the end pointer already points to the list’s tail, it wraps around to the head to accommodate the recycled pages.

\myparagraph{Generation Phase}
At each generation step, a head allocates a new page only if either its high-precision or low-precision pages are full, requiring at most one additional page per step.
Each head independently checks its page availability and, if needed, allocates a new page in parallel using the prefix-sum-based approach.
Unlike in the prompt phase, page recycling is not performed during generation, as the total number of stored tokens either remains the same if an old token is evicted or increases by one if no eviction occurs.
Once a request is finished, all pages allocated for that request are recycled, freeing up memory for incoming requests.

\myparagraph{Supporting Additional Precision Levels}
The proposed memory management system naturally extends to support more than two precision levels, by incorporating additional page tables. For example, accommodating three precision levels can be achieved by combining a unidirectional page table with a bidirectional one. Similarly, four precision levels can be supported using two bidirectional page tables. This composable design preserves memory efficiency while providing the flexibility needed to handle increasingly fine-grained precision differentiation.

\vspace{-0.2em}

\section{Implementation}
\label{sec:impl}

We implement \system on top of vLLM, comprising 4.5K lines of CUDA/C++ code and 9K lines of Python.
We first outline the architecture of \system and then detail our custom GPU attention kernel, designed to efficiently support the differentiated KV cache compression.

\vspace{-0.3em}

\subsection{\system Architecture}

\begin{figure}
\centering
\includegraphics[trim={0cm 7.0cm 0cm 0cm},clip, width=1.0\linewidth]{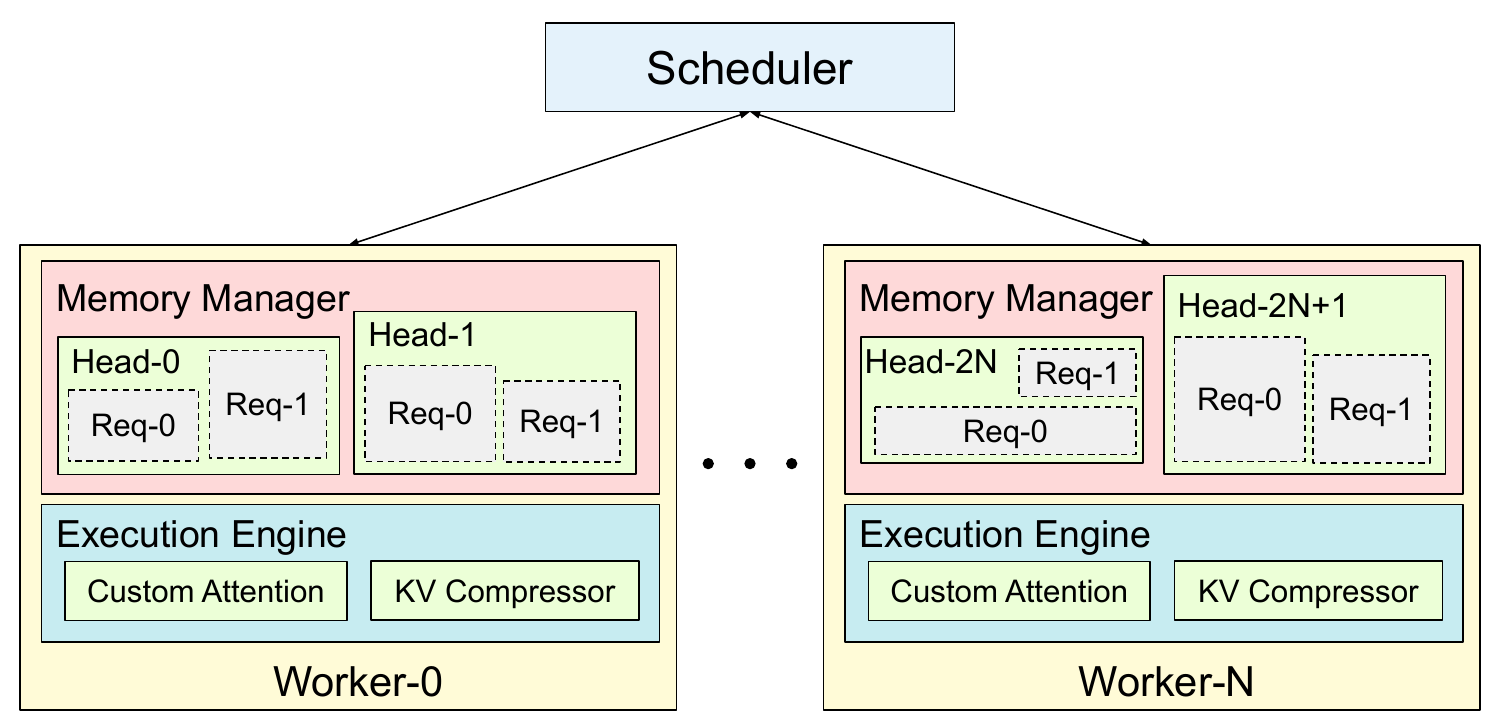}
\caption{\system architecture. }
\vspace{-0.23in}
\label{fig:compactkv_arch}
\end{figure}

The architecture of \system is illustrated in Figure~\ref{fig:compactkv_arch}. At each inference step, the scheduler batches as many requests as possible within the available GPU memory to maximize throughput, sending the selected requests to all workers. Each GPU hosts one worker, responsible for executing a partition of the model~\cite{shoeybi2019megatron, huang2019gpipe, narayanan2019pipedream, narayanan2021efficient}. 
To facilitate efficient execution, each worker includes a dedicated memory manager that oversees the KV cache for its assigned attention heads, as outlined in Section~\ref{sec:system_design}, and an execution engine for model computation.
To support differentiated KV cache compression, the execution engine integrates a KV compressor and a custom GPU attention kernel. After computing key and value vectors, the KV compressor is invoked to compress them following the policy described in Section~\ref{sec:unified_kv_compression}, storing the results in the KV cache. The custom GPU attention kernel then efficiently computes the attention output using the compressed KV cache.


\vspace{-0.1in}

\subsection{Efficient Attention Kernel}

We develop a custom attention kernel that efficiently supports differentiated KV cache compression, leveraging reduced memory access volume to accelerate performance.
Notably, the kernel design is portable across GPUs and other NPUs, as mainstream architectures support vectorized memory access and layout-aware tiling~\cite{jouppi2017datacenter,zuo2025serving}.
We detail the GPU implementation in the following paragraphs.

Overall, we assign each thread block to process a single attention head per sequence.
To mitigate the potential load imbalance caused by mixed-precision quantization, we let thread warps iterate over high-precision pages first, followed by low-precision pages.
Each page is processed in two phases: dot product between the query and keys to derive attention scores, and weighted sum of values.
These two phases follow different computation patterns: reduction over the feature dimension and token dimension, respectively.
To ensure coalesced and vectorized memory accesses for both phases, we design tailored data layouts and parallelization strategies, elaborated as follows.

For key processing, each warp handles one page at a time, with threads divided into groups responsible for distinct key vectors.
Within a group, each thread fetches its assigned key elements in a vectorized manner, performs dequantization on the fly, and computes a partial dot product between the key elements and corresponding query elements.
Using a straightforward layout such as [$F$, $N_{\text{tokens}}$] for keys, where $F$ denotes the feature dimension length, would cause threads within a group to fetch non-contiguous elements strided by $N_{\text{tokens}}$ when parallelizing across the feature dimension.
Alternatively, using [$ N_{\text{tokens}}$, $F$] would cause thread groups to fetch non-contiguous elements strided by $F$ when parallelizing across tokens.
Both layouts suffer inefficient strided memory accesses, leading to low bandwidth utilization.
Instead, we organize the layout of keys as [
$\frac{F}{K_{\text{vec}} \times K_{\text{group}}}$,
$N_{\text{tokens}}$,
$K_{\text{group}}$,
$K_{\text{vec}}$], where $K_{\text{vec}}$ denotes the vectorization factor and $K_{\text{group}}$ denotes the number of threads per group.
During each execution step, each thread fetches $K_{\text{vec}}$ consecutive elements, threads within a group collectively fetch $K_{\text{group}}$ adjacent chunks, and groups within the warp access contiguous chunks along the $N_{\text{tokens}}$ dimension.
As a result, the combined memory accesses of all threads in a warp are contiguous, enabling memory coalescing and maximizing bandwidth utilization.

For value processing, tokens in a page are evenly distributed across thread groups in a warp; within each group, individual threads perform sum reductions over the feature dimension and save their accumulation results in registers.
Once all groups finish, a tree reduction aggregates these partial results and produces the output.
The number of registers required per thread is proportional to the feature dimension range assigned to it, while during key processing each thread only requires a single register to store the partial dot product.
Consequently, vectorization along the feature dimension, as used in key processing, is not suitable for value processing, because it would significantly increase register pressure by forcing each thread to handle a larger portion of the feature dimension.
To better align with the computation pattern of reduction across tokens, we apply vectorization to the token dimension instead, which reduces register pressure and ensures more effective parallelization.
We organize the layout of values as [
$\frac{F}{V_{\text{group}}}$, $\frac{N_{\text{tokens}}}{V_{\text{vec}}}$, $V_{\text{group}}$,
$V_{\text{vec}}$] accordingly.

Additionally, for ultra-long sequences, the attention kernel supports parallelization along the sequence dimension.
It splits the sequence into multiple segments, processes them in parallel using the aforementioned method within separate thread blocks, and then merges the results with minimal computation overhead.

\vspace{-0.1in}

\section{Evaluation}

\vspace{-0.08in}


We first evaluate the effectiveness of \system's differentiated KV cache compression policy.
Next, we evaluate the efficiency of the memory manager as well as the end-to-end throughput improvements achieved by \system.

\vspace{-0.1in}

\subsection{Experiment Setup}
\label{sec:experiment_setup}

We evaluate \system on several models spanning three major families: \llama3-8B and 70B~\cite{touvron2023llama}, Qwen2.5-7B and 32B~\cite{hui2024qwen2}, and the recent thinking models QwQ-32B~\cite{qwq32b}, R1-Distill-Qwen-14B and R1-Distill-Llama-8B~\cite{guo2025deepseek}, which generate extended chains of thought and exhibit strong capabilities on complex reasoning tasks.

We select benchmarks that align with those commonly used in recent LLM technical reports~\cite{dubey2024llama,hui2024qwen2,guo2024deepseek,liu2024deepseek,qwq32b}, covering a diverse set of task domains, including general knowledge (MMLU~\cite{hendrycks2020measuring} and MMLU-Pro~\cite{taghanaki2024mmlu}), mathematics (GSM8K~\cite{cobbe2021training} and MATH~\cite{lewkowycz2022solving}), code generation (HumanEval+~\cite{chen2021evaluating, evalplus} and MBPP+~\cite{austin2021program, evalplus}), and long-context understanding (LongBench~\cite{bai2023longbench}).
In addition, we assess \system's generation quality on thinking models using two particularly challenging benchmarks: AIME24~\cite{aime24}, a mathematics competition dataset, and GPQA~\cite{rein2024gpqa}, which evaluates graduate-level science reasoning.
We measure the throughput and latency on NVIDIA L40 GPUs, each with 48 GB of memory~\cite{L40}.
In all experiments, model weights are stored in FP16 precision.

\vspace{-0.1in}


\subsection{Differentiated KV Compression Policy}
\label{sec:eval-accuracy}


We first assess the effectiveness of differentiated KV quantization and dynamic sparsity individually, and then evaluate the combined benefits of the proposed differentiated KV cache compression.
To ensure the reliability of our results and mitigate the influence of noise and randomness inherent in floating-point computations, each experiment is repeated five times with the dataset randomly shuffled, and the reported results represent the average across these runs.

\begin{figure}[t]
\centering
\includegraphics[width=1.0\linewidth]{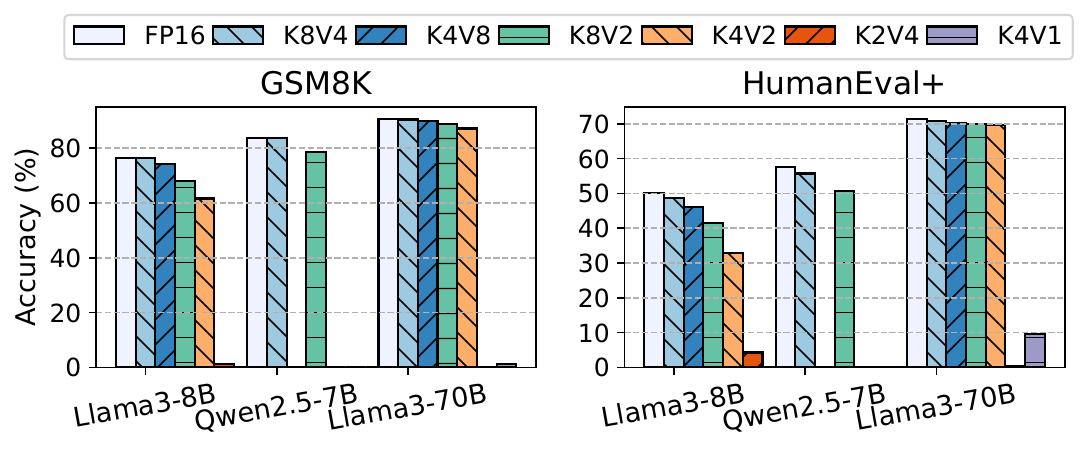}
\vspace{-1.8em}
\caption{Accuracy of differentiated KV quantization.}
\label{fig:heterokv}
\vspace{-0.1in}
\end{figure}

\myparagraph{Evaluating Differentiated KV Quantization}
Figure~\ref{fig:heterokv} shows the model generation quality of differentiated KV quantization on GSM8K and HumanEval+, namely K8V4 and K4V2, compared to FP16 baselines, across \llama3-8B \& 70B and Qwen2.5-7B.
To validate our intuition that keys have a more significant impact than values (Section~\ref{sec:impact_kv}), we additionally evaluate the mirror configurations, namely K4V8 and K2V4, where values are stored with higher precision than keys, as well as more skewed variants, K8V2 and K4V1.

Results show that K8V4 matches the accuracy of the FP16 baseline across all models and benchmarks. In contrast, its mirror configuration, K4V8, exhibits noticeable accuracy degradation, particularly for Qwen2.5-7B, where accuracy drops to nearly zero on both tasks. This pronounced sensitivity to 4-bit key quantization in Qwen2.5-7B is likely due to its GQA architecture, which applies aggressive KV compression with a queries-per-KV ratio of 7, substantially higher than the ratio of 4 used in \llama3-8B.
Similarly, K4V2 retains over 65\% of FP16 accuracy across both benchmarks on \llama3-8B and 70B, while the mirror configuration K2V4 results in near-zero accuracy. 
The more skewed variant K8V2 retains at least 83\% of FP16 accuracy across all models, even for Qwen2.5-7B, reinforcing the importance of assigning higher precision to keys. 
On the other hand, K4V1 yields almost zero accuracy, suggesting that 2 bits is the lower bound for effective value quantization.
In summary, these results confirm that keys play a more critical role than values, and demonstrate the effectiveness of differentiated KV quantization in preserving accuracy while enabling aggressive compression.

The accuracy trends across various differentiated KV quantization configurations motivate the design of adopting two precision levels: K8V4 and K4V2. K8V4 matches the generation quality of the FP16 baseline and is applied to important tokens to preserve model accuracy. K8V2 and K4V2, when applied uniformly, both lead to noticeable accuracy degradation. Among them, K4V2 offers better efficiency and is thus selected for compressing less significant tokens, enabling opportunistic KV cache savings with minimal quality loss.
As we demonstrate later, the K8V4–K4V2 scheme achieves near-lossless generation quality across multiple models and benchmarks, obviating the need for an intermediate precision level like K8V2. Likewise, introducing an even lower precision level such as K4V1 is suboptimal: not only does K4V1 significantly degrade accuracy, but the additional memory savings from a three-level configuration K8V4–K4V2–K4V1 are marginal compared to the K8V4–K4V2 design.
That said, future models employing more aggressive training-time KV compression (e.g., architectures with higher queries-per-KV ratios in GQA) may benefit from an additional high-precision level (e.g., FP16–K8V4 or FP16–K8V4–K4V2), which we leave to future investigation.

\begin{figure}[t]
\centering
\includegraphics[width=1.0\linewidth]{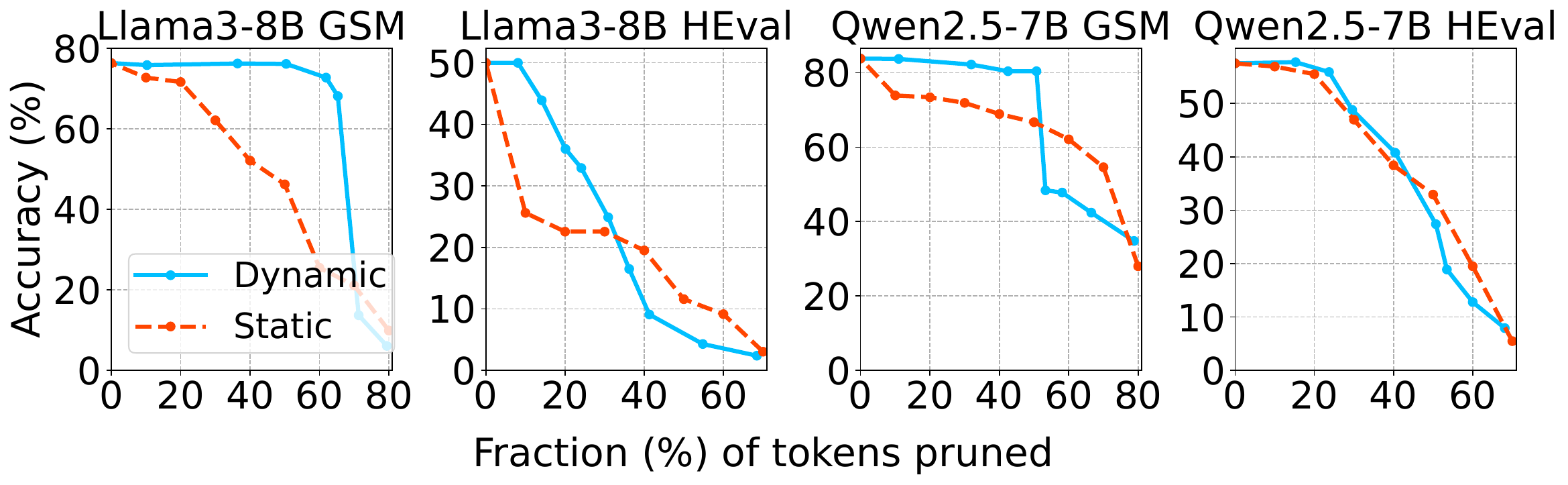}
\vspace{-2em}
\caption{Accuracy of dynamic vs. static sparsity.}
\label{fig:dynamic_sparsity}
\vspace{-0.2in}
\end{figure}

\myparagraph{Evaluating Dynamic Sparsity}
We evaluate the effectiveness of per-head dynamic sparsity (Section~\ref{sec:dynamic_sparsity}) in identifying critical tokens, comparing it to the static sparsity method used in SnapKV~\cite{li2024snapkv}, which allocates an equal memory budget to all attention heads.
Figure~\ref{fig:dynamic_sparsity} reports results on Qwen2.5-7B and \llama3-8B across GSM8K and HumanEval+. The x-axis indicates the percentage of tokens pruned, while the y-axis shows the resulting task accuracy.
Dynamic sparsity significantly outperforms static sparsity in \llama3-8B, maintaining full accuracy with 50\% tokens pruned on GSM8K, and 10\% tokens pruned for the more sensitive HumanEval+. For Qwen2.5-7B, dynamic sparsity also consistently surpasses static sparsity, achieving accuracy higher than 80\% on GSM8K with 50\% tokens pruned. While the improvement is less pronounced on HumanEval+, dynamic sparsity still yields higher accuracy at equivalent pruning ratios. 
In summary, per-head dynamic sparsity is superior to static sparsity by leveraging head-specific significance. 

\begin{figure}[t]
\centering
\includegraphics[width=1.0\linewidth]{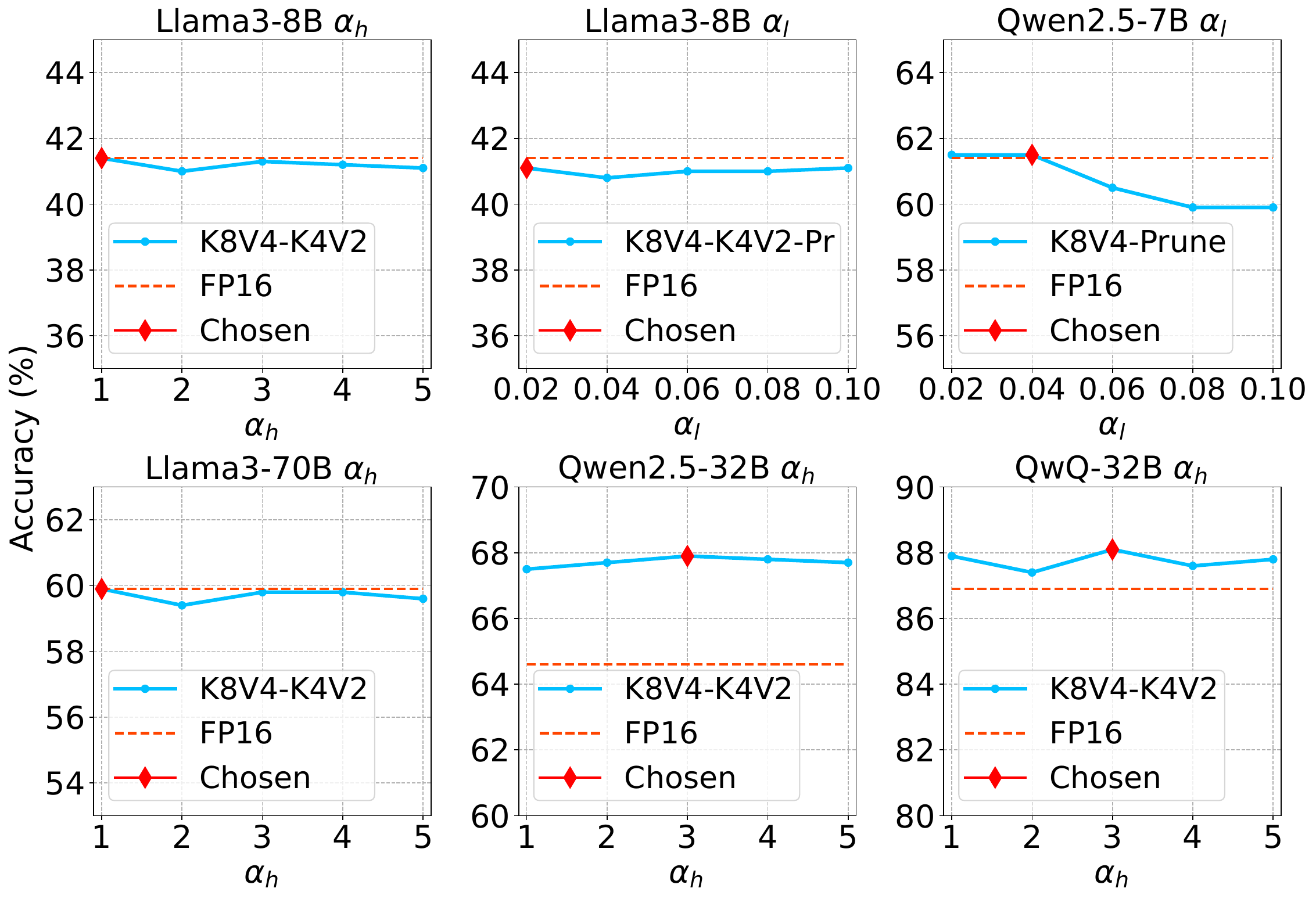}
\vspace{-1.8em}
\caption{Calibrating the high- and low-precision thresholds $\alpha_h$ and $\alpha_l$ on the training split of the MATH dataset.}
\label{fig:thresh_calibration}
\vspace{-1.8em}
\end{figure}

\myparagraph{Parameter Calibration}
As described in Section~\ref{sec:unified_kv_compression}, \system compares the significance score of a token with its theoretical average, $\frac{1}{N}$, where $N$ denotes the sequence length. A token is quantized to high precision if its score exceeds $\frac{\alpha_h}{N}$, to low precision if its score falls within the interval $[\frac{\alpha_l}{N}, \frac{\alpha_h}{N}]$, and is pruned otherwise.
To calibrate the appropriate parameters $\alpha_h$ and $\alpha_l$, we utilize the training split of MATH~\cite{lewkowycz2022solving}.
We select MATH as the calibration dataset as it is comprised of complex mathematical reasoning tasks with high text information density, where even the removal of a single symbol can invalidate an argument. This makes MATH highly suitable for parameter fitting, as the resulting compression parameters are expected to generalize well across other tasks. Additionally, MATH provides a dedicated training split, ensuring that parameter fitting does not involve testing data and mitigating potential overfitting.

We profile a consistent range of parameters across all evaluated models, as depicted in Figure~\ref{fig:thresh_calibration}. The x-axis represents the profiled parameter values, while the y-axis shows the resulting accuracy on the calibration dataset.
Specifically, we profile $\alpha_h$ within the integer range $[1, 5]$, progressively relaxing the high-precision threshold beyond the theoretical average. This adjustment accounts for aggregating scores from multiple attention heads by maximum for GQA, as mentioned in Section~\ref{sec:unified_kv_compression}. The profiled values of $\alpha_h$ achieve FP16-equivalent accuracy across all models, except for Qwen2.5-7B, which exhibits heightened sensitivity to 4-bit key quantization.
Accordingly, we select the values of $\alpha_h$ that yield the highest accuracy based on profiling results: $\alpha_h = 1$ for \llama3-8B and 70B, and $\alpha_h = 3$ for Qwen2.5-32B and QwQ-32B, and disable low‑precision quantization for Qwen2.5‑7B.
Subsequently, we profile the low-precision threshold $\alpha_l$, while keeping $\alpha_h$ fixed. We observe that setting $\alpha_l = 0.1$ induces at least 5\% accuracy degradation on the larger models. As a result, we profile five equally spaced values within the range $[0, 0.1]$ and select the one yielding the highest accuracy for each model. The final selected values of $\alpha_l$ are $0.02$ for \llama3-8B, $0.04$ for Qwen2.5-7B, and $0$ for the remaining models. 
While the profiled values cover a limited portion of the entire parameter space, as will be demonstrated in the following sections, Diff-KV significantly outperforms existing baselines with these fitted parameters, highlighting its potential effectiveness.

\begin{figure}[t]
\centering
\includegraphics[width=1.0\linewidth]{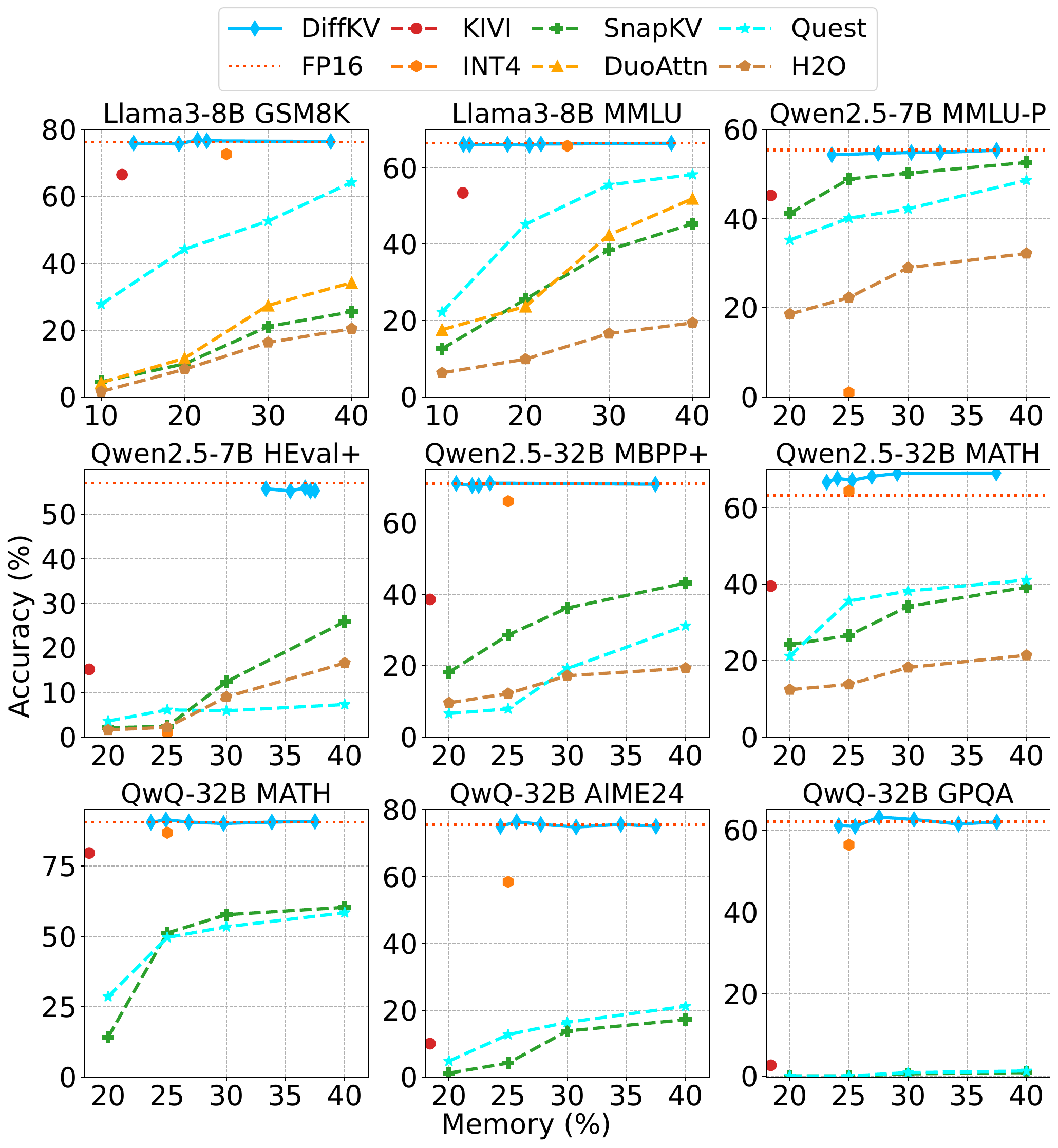}
\vspace{-2em}
\caption{KV cache memory normalized to vLLM vs. benchmark accuracy tradeoff of \system.}
\label{fig:pareto}
\vspace{-1em}
\end{figure}

\myparagraph{Evaluating Differentiated Compression Policy}
Finally, we evaluate the combined benefits of \system's KV cache compression policy. We compare \system against several state-of-the-art baselines, including pruning-based methods such as H2O~\cite{zhang2024h2o}, SnapKV~\cite{li2024snapkv}, and DuoAttention~\cite{xiao2024duoattention}, quantization-based approaches such as 4-bit KV, KIVI~\cite{liu2024kivi}, and QAQ~\cite{dong2024qaq}, as well as KV partial loading techniques like Quest~\cite{tang2024quest}.
Table~\ref{tab:accuracy_and_memory_usage} compares the accuracy and KV cache memory usage of \system and the best-performing baselines, both normalized to vLLM~\cite{kwon2023efficient}, for the evaluated non-thinking models. The parameters of \system are tuned as described previously. For the pruning-based baselines and Quest, we tune their memory usage to 50\% of the baseline.
\system achieves near-lossless accuracy relative to the FP16 baseline, with an average degradation of only $0.3\%$, while using $19.3\%$ to $36.7\%$ memory. In contrast, all other baselines incur more significant accuracy degradation at comparable or higher memory usage. Furthermore, \system dynamically adapts its memory usage based on the task's information density. For example, \system allocates less memory for the 5-shot MMLU benchmark compared to the 0-shot HumanEval+, where the prompt consists solely of the function name and comments.

\system also outperforms baseline methods in long-context scenarios, as shown in Table~\ref{tab:longbench_eval}, where we present results from one benchmark in each category of LongBench. For \llama3-8B and Qwen2.5-7B, \system achieves memory usage of $14.9\%$ and $27.0\%$, respectively, while baseline methods are tuned to $25\%$ memory usage. Notably, \system consistently achieves superior accuracy across all categories, with an average degradation of only $0.4\%$, while utilizing less or comparable memory compared to the baseline methods.

Furthermore, the generation quality gap between \system and baseline methods becomes even more pronounced for the thinking models 
QwQ-32B, R1-Distill-Qwen-14B and R1-Distill-Llama-8B on the more challenging benchmarks, as demonstrated in Table~\ref{tab:qwq_eval}.
\system achieves FP16-comparable generation quality using $27.4\%$, $29.4\%$ and $23.5\%$ of the memory on average for QwQ-32B, R1-Distill-Qwen-14B and R1-Distill-Llama-8B, respectively, while all other methods experience significant generation quality degradation. On the math competition benchmark AIME24, the most accurate baseline, 4-bit KV quantization incurs a $16.0\%$ accuracy degradation. On the graduate-level science benchmark GPQA, the accuracy of pruning-based and KV partial loading methods drops to almost zero, while using $50\%$ of the full KV cache.
The larger gap between \system and prior work in long CoT generation
highlights the increased challenges such tasks pose for KV cache compression. In long CoT generation, tokens are generated autoregressively based on previous tokens, and errors introduced by compression are accumulated and propagated across generation steps. In contrast, in long prompt scenarios, the majority of the text is provided as ground truth in the prompt, and the model-generated portion is significantly shorter, limiting the impact of errors introduced by KV cache compression.

Lastly, Figure~\ref{fig:pareto} depicts the accuracy-memory trade-offs of \system under the profiled parameter values, namely $\alpha_h \in [0, 5]$  (step size of 1) for all models and $\alpha_l \in [0.02, 0.1]$ (step size of 0.02) for \llama3-8B and Qwen2.5-7B.
We also include the accuracy of baseline methods within similar memory usage ranges for comparison. \system consistently matches the FP16 baseline accuracy across different models and benchmarks within the profiled range of memory usage, demonstrating its robustness. In contrast, baseline methods experience significant accuracy degradation, particularly pruning-based and KV partial loading methods, which suffer rapid accuracy declines as memory usage decreases.
In summary, \system provides a superior accuracy-memory tradeoff compared to state-of-the-art methods.

\aboverulesep=0ex
\belowrulesep=0ex
\renewcommand{\arraystretch}{1.15}

\begin{table*}[h!]
\centering
\caption{Accuracy and memory usage of DiffKV and the best-performing baseline methods across models and benchmarks. Numbers in parentheses represent memory usage normalized to FP16. The memory usage is 50.0\% for SnapKV, Quest and DuoAttn, 25.0\% for INT4, 18.8\% for QAQ, and 12.5\% for KIVI. \system achieves accuracy close to FP16, consistently outperforming prior quantization and pruning approaches.}
\vspace{-1em}
\begin{adjustbox}{width=1.0\linewidth}
\large
\begin{tabular}{c|ccccc|ccccc|ccccc|ccc}
\toprule
\multirow{2}{*}{\Large Benchmarks} & \multicolumn{5}{c|}{\Large Llama3-8B} & \multicolumn{5}{c|}{\Large Qwen2.5-7B} & \multicolumn{5}{c|}{\Large Qwen2.5-32B} & \multicolumn{3}{c}{\Large Llama3-70B} \\
\multicolumn{1}{c|}{} & \cellcolor{cyan!15}\textbf{FP16} & \cellcolor{red!25}\textbf{DiffKV} & \textbf{INT4} & \textbf{QAQ} & \textbf{DuoAttn} & \cellcolor{cyan!15}\textbf{FP16} & \cellcolor{red!25}\textbf{DiffKV} & \textbf{Quest} & \textbf{SnapKV} & \textbf{KIVI} & \cellcolor{cyan!15}\textbf{FP16} & \cellcolor{red!25}\textbf{DiffKV} & \textbf{Quest} & \textbf{INT4} & \textbf{KIVI} & \cellcolor{cyan!15}\textbf{FP16} & \cellcolor{red!25}\textbf{DiffKV} & \textbf{INT4} \\
\midrule
\textbf{GSM8K}      & \cellcolor{cyan!15}76.3 & \cellcolor{red!25}75.7 (19.3\%) & 72.6 & 71.7 & 32.2 & \cellcolor{cyan!15}83.5 & \cellcolor{red!25}83.6 (26.8\%) & 76.7 & 66.7 & 72.2 & \cellcolor{cyan!15}90.4 & \cellcolor{red!25}90.2 (23.6\%) & 82.8 & 88.7 & 79.3 & \cellcolor{cyan!15}90.5 & \cellcolor{red!25}90.2 (21.6\%) & 89.3 \\
\textbf{MATH}       & \cellcolor{cyan!15}28.1 & \cellcolor{red!25}27.9 (21.6\%) & 23.9 & 23.5 & 9.3  & \cellcolor{cyan!15}58.0 & \cellcolor{red!25}57.7 (32.3\%) & 41.7 & 45.1 & 39.5 & \cellcolor{cyan!15}63.2 &  \cellcolor{red!25}67.2 (25.3\%)  & 44.3 & 64.7 & 42.6 & \cellcolor{cyan!15}48.7 & \cellcolor{red!25}47.6 (21.3\%) & 45.3 \\
\textbf{MMLU}       & \cellcolor{cyan!15}66.5 & \cellcolor{red!25}66.1 (17.8\%) & 65.2 & 60.9 & 42.4 & \cellcolor{cyan!15}75.1 & \cellcolor{red!25}74.7 (30.3\%) & 64.7 & 48.2 & 53.1 & \cellcolor{cyan!15}83.8 & \cellcolor{red!25}83.8 (23.1\%)  & 68.5 & 83.4 & 70.5 & \cellcolor{cyan!15}81.0 & \cellcolor{red!25}80.9 (20.1\%) & 79.4 \\
\textbf{MMLU-Pro}   & \cellcolor{cyan!15}41.5 & \cellcolor{red!25}41.0 (21.5\%) & 39.1 & 38.2 & 34.6 & \cellcolor{cyan!15}55.4 & \cellcolor{red!25}54.9 (30.2\%) & 51.6 & 49.7 & 45.2 & \cellcolor{cyan!15}67.8 & \cellcolor{red!25}67.4 (24.5\%)  & 59.5 & 66.2 & 54.5 & \cellcolor{cyan!15}60.1 & \cellcolor{red!25}60.0 (21.5\%) & 59.2 \\
\textbf{HumanEval+} & \cellcolor{cyan!15}50.0 & \cellcolor{red!25}48.0 (27.6\%) & 45.1 & 7.9 & 4.8 & \cellcolor{cyan!15}57.5 & \cellcolor{red!25}55.9 (36.7\%) & 8.5  & 32.9 & 15.2 & \cellcolor{cyan!15}49.4 & \cellcolor{red!25}49.4 (26.2\%)  & 11.0 & 45.8 & 15.3 & \cellcolor{cyan!15}71.3 & \cellcolor{red!25}71.5 (26.7\%) & 70.1 \\
\textbf{MBPP+}      & \cellcolor{cyan!15}59.3 & \cellcolor{red!25}61.6 (17.6\%) & 58.8 & 56.3 & 35.6 & \cellcolor{cyan!15}64.3 & \cellcolor{red!25}62.8 (27.9\%) & 33.3 & 44.7 & 18.3 & \cellcolor{cyan!15}71.1 & \cellcolor{red!25}70.5 (21.8\%)   & 35.2  & 67.3 & 38.6 & \cellcolor{cyan!15}68.6 & \cellcolor{red!25}69.7 (22.0\%) & 67.2 \\
\bottomrule
\end{tabular}
\end{adjustbox}
\vspace{-0.15in}
\label{tab:accuracy_and_memory_usage}
\end{table*}

\begin{table}[h!]
\centering
\caption{Evaluating DiffKV on LongBench. DiffKV uses 14.9\% memory for Llama3.1-8B and 27.0\% for Qwen2.5-7B. For Quest and SnapKV, the memory usage is 25\%.}
\vspace{-1em}
\begin{adjustbox}{width=1\linewidth}
\large
\begin{tabular}{c|c|cccccc}
\toprule
\multicolumn{2}{c|}{} & Qasper & HotpotQA & GovReport & TREC & PCount & Lcc \\
\midrule
\multicolumn{1}{c|}{\multirow{4}{*}{\textbf{Llama3.1-8B}}} & \cellcolor{cyan!15}\textbf{FP16} & \cellcolor{cyan!15}40.9 & \cellcolor{cyan!15}61.3 & \cellcolor{cyan!15}34.0 & \cellcolor{cyan!15}73.0 & \cellcolor{cyan!15}6.9 & \cellcolor{cyan!15}62.2 \\
& \cellcolor{red!25}\textbf{DiffKV} & \cellcolor{red!25}42.8 & \cellcolor{red!25}61.3 & \cellcolor{red!25}32.9 & \cellcolor{red!25}73.0 & \cellcolor{red!25}7.2 & \cellcolor{red!25}61.9 \\
& \textbf{Quest} & 38.9 & 59.8 & 30.6 & 66.5 & 6.4 & 56.8 \\
& \textbf{SnapKV} & 39.2 & 59.6 & 30.1 & 65.9 & 6.2 & 57.2 \\
\midrule
\multicolumn{1}{c|}{\multirow{4}{*}{\textbf{Qwen2.5-7B}}} & \cellcolor{cyan!15}\textbf{FP16} & \cellcolor{cyan!15}26.5 & \cellcolor{cyan!15}27.8 & \cellcolor{cyan!15}33.4 & \cellcolor{cyan!15}71.0 & \cellcolor{cyan!15}5.7 & \cellcolor{cyan!15}61.9 \\
& \cellcolor{red!25}\textbf{DiffKV} & \cellcolor{red!25}26.4 & \cellcolor{red!25}28.2 & \cellcolor{red!25}32.2 & \cellcolor{red!25}70.0 & \cellcolor{red!25}5.3 & \cellcolor{red!25}62.3 \\
& \textbf{Quest} & 23.6 & 25.2 & 31.4 & 65.6 & 4.3 & 53.3 \\
& \textbf{SnapKV} & 22.6 & 25.5 & 30.8 & 64.6 & 4.5 & 52.6 \\
\bottomrule
\end{tabular}
\end{adjustbox}
\label{tab:longbench_eval}
\end{table}


\begin{table}[h!]
\centering
\caption{Evaluating DiffKV on thinking models.}
\vspace{-1em}
\begin{adjustbox}{width=1\linewidth}
\begin{tabular}{c|c|cccccc}
\toprule
\multicolumn{2}{c|}{} & \cellcolor{cyan!15}\textbf{FP16} & \cellcolor{red!25}\textbf{DiffKV} & \textbf{INT4} & \textbf{KIVI} & \textbf{Quest}  & \textbf{SnapKV} \\
\midrule
\multicolumn{1}{c|}{\multirow{3}{*}{\textbf{QwQ-32B}}} & \textbf{MATH} & \cellcolor{cyan!15}90.6 & \cellcolor{red!25}90.6 (26.8\%) & 87.3 & 80.6 & 63.3 & 65.4\\
& \textbf{GPQA} & \cellcolor{cyan!15}62.1 & \cellcolor{red!25}63.2 (27.6\%) & 57.8 & 2.6  & 1.2  & 0.8 \\
& \textbf{AIME24} & \cellcolor{cyan!15}75.5 & \cellcolor{red!25}75.3 (27.8\%) & 64.3 & 10.0 & 23.3 & 19.2 \\
\midrule
\multicolumn{1}{c|}{\multirow{3}{*}{\parbox{1.8cm}{\textbf{R1-Distill-}\\ \textbf{Qwen-14B}}}} & \textbf{MATH} & \cellcolor{cyan!15}94.2 & \cellcolor{red!25}94.1 (29.5\%) & 92.7 & 69.2 & 68.4 & 62.6 \\
& \textbf{GPQA} & \cellcolor{cyan!15}55.7 & \cellcolor{red!25}55.6 (29.2\%) & 52.0 & 0.0 & 0.5 & 0.0 \\
& \textbf{AIME24} & \cellcolor{cyan!15}67.0 & \cellcolor{red!25}67.0 (29.6\%) & 61.6 & 10.0 & 20.2 & 17.2 \\
\midrule
\multicolumn{1}{c|}{\multirow{3}{*}{\parbox{1.8cm}{\textbf{R1-Distill-}\\ \textbf{Llama-8B}}}} & \textbf{MATH} & \cellcolor{cyan!15}88.8 & \cellcolor{red!25}88.7 (23.2\%) & 86.1 & 48.2 & 58.9 & 54.2 \\
& \textbf{GPQA} & \cellcolor{cyan!15}47.4 & \cellcolor{red!25}46.5 (23.5\%) & 42.9 & 0.0 & 0.5 & 0.5 \\
& \textbf{AIME24} & \cellcolor{cyan!15}51.0 & \cellcolor{red!25}51.0 (23.8\%) & 42.3 & 6.6 & 18.2 & 15.4 \\
\bottomrule
\end{tabular}
\end{adjustbox}
\vspace{-0.15in}
\label{tab:qwq_eval}
\end{table}

\begin{figure}[t]
\centering
\includegraphics[width=1\linewidth]{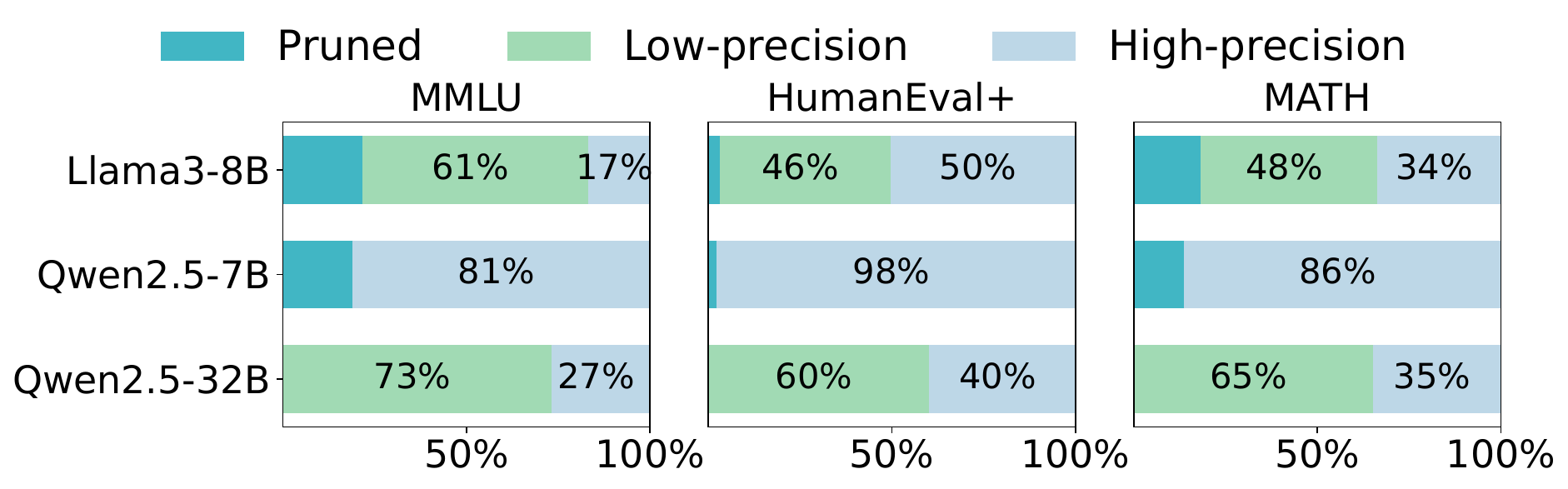}
\vspace{-1.2em}
\caption{Fraction of tokens pruned, quantized to low precision, and quantized to high precision across different benchmarks and models.}
\label{fig:compression_breakdown}
\vspace{-0.5em}
\end{figure}

\myparagraph{KV Compression Breakdown}
Figure~\ref{fig:compression_breakdown} shows the fraction of tokens that are pruned, quantized to low precision, and quantized to high precision across three benchmarks (MMLU, HumanEval+, and MATH) for three different models. Two key observations emerge.
First, sparsity levels vary substantially across workloads. The general knowledge benchmark MMLU exhibits the highest sparsity, likely due to the lower information density of natural language and the use of 5-shot prompts. In contrast, the reasoning-intensive benchmarks MATH and HumanEval+ display lower sparsity, with HumanEval+ being the least sparse, partly due to its 0-shot setting.
Second, despite being calibrated on challenging reasoning-heavy workloads, \system's KV compression policy adapts effectively to the varying sparsity patterns across different tasks, highlighting its generalizability to diverse scenarios.

\vspace{-0.1in}

\subsection{System Performance}
\label{sec:eval-perf}

In this section, we provide a comprehensive evaluation of \system's performance, comparing it against state-of-the-art systems including vLLM, Quest, and SnapKV across various models.
For models that fit within the memory of a single GPU, such as Llama3-8B and Qwen2.5-7B, we use one GPU.
For larger models, we parallelize the computation across multiple GPUs: four GPUs for Llama3-70B, and two GPUs for Qwen2.5-32B and QwQ-32B.
Our evaluation focuses on latency breakdown, attention kernel speedup, and end-to-end throughput to demonstrate the efficiency and scalability of \system.

\begin{figure}[t]
    \begin{subfigure}{0.49\linewidth}
        \centering
        \includegraphics[width=\linewidth]{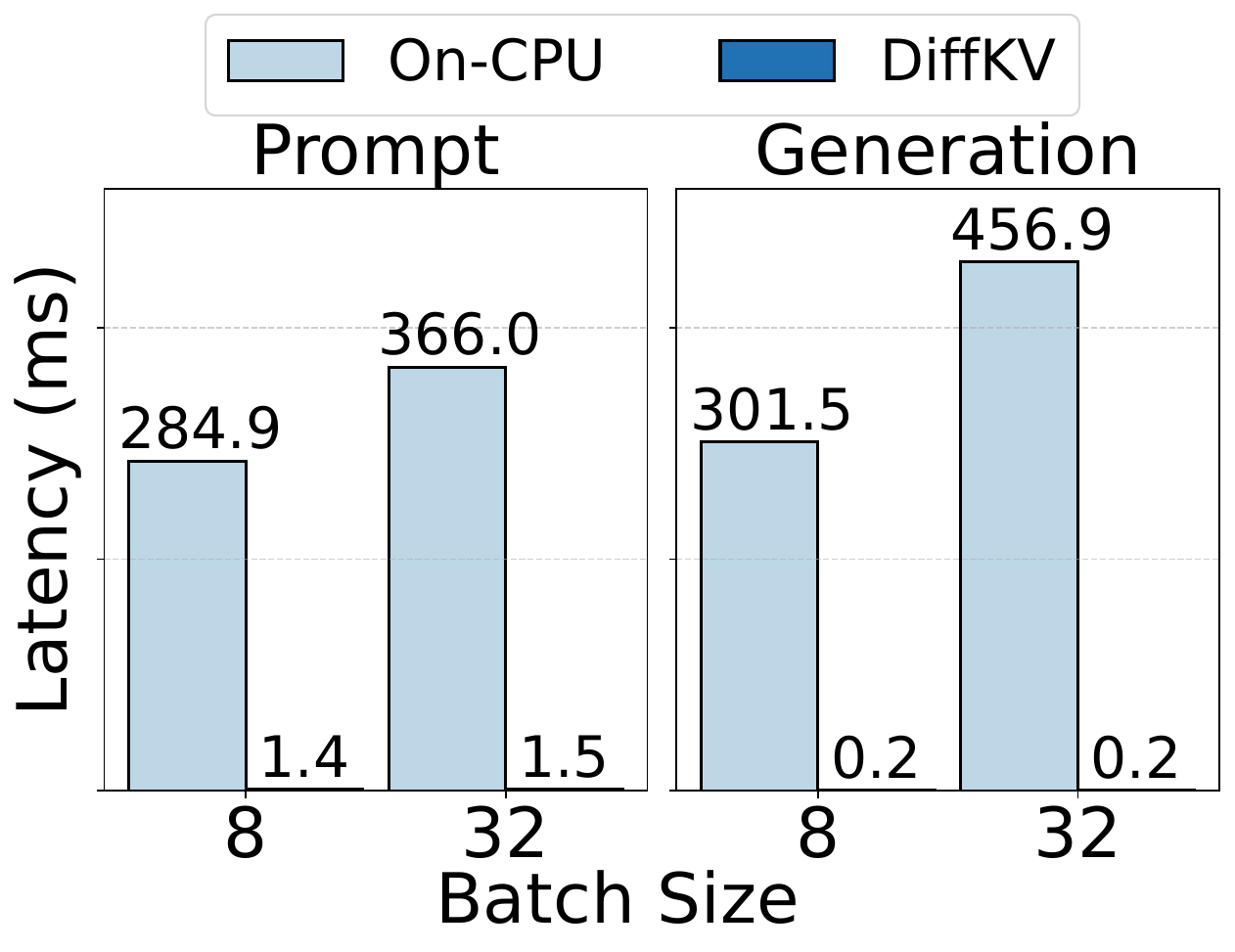}
        \vspace{-2em}
        \caption{Memory management}
        \label{fig:latency_memory_management}
    \end{subfigure}
    \hfill
    \centering
    \begin{subfigure}{0.49\linewidth}
        \centering
        \includegraphics[width=\linewidth]{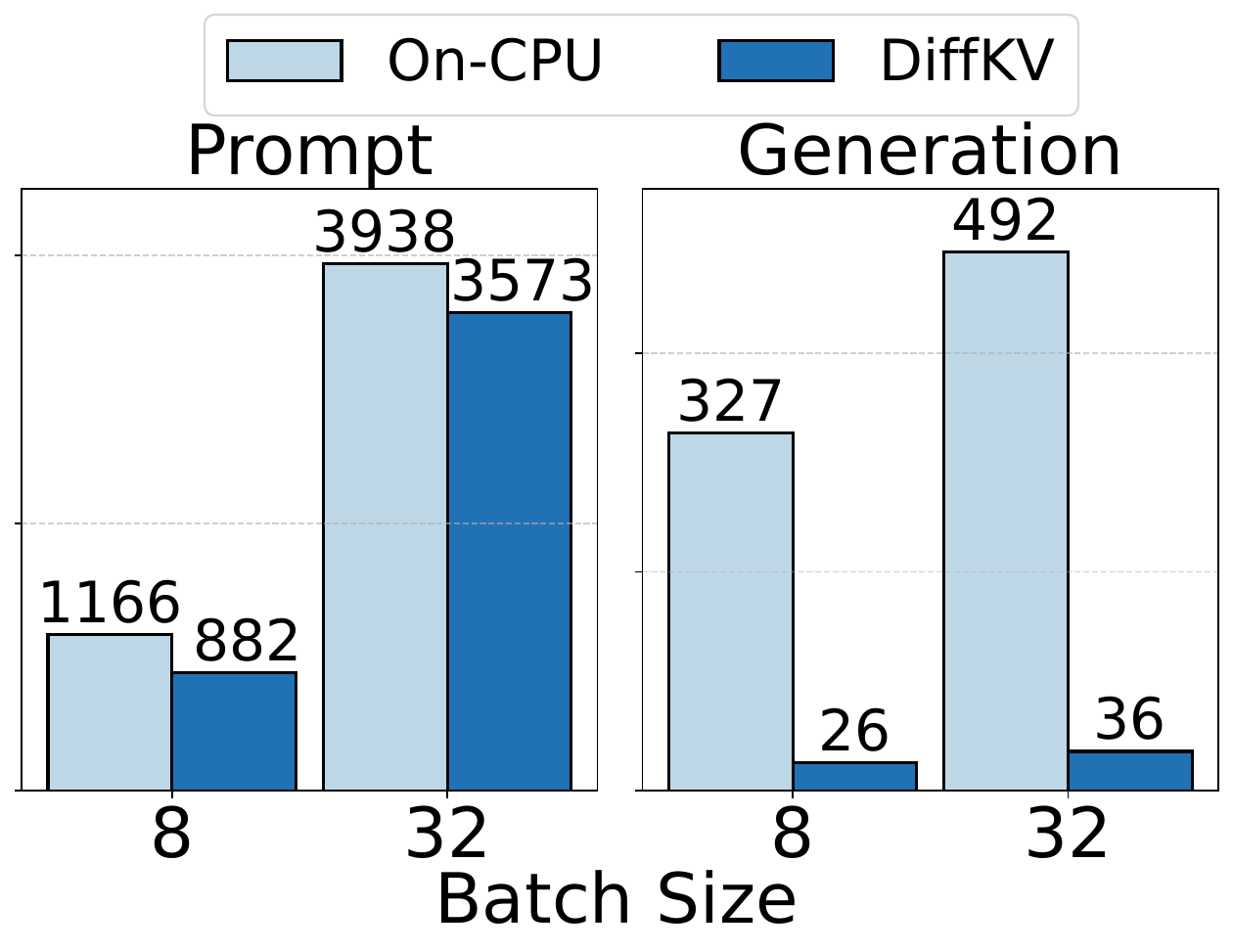}
        \vspace{-2em}
        \caption{One entire inference step}
        \label{fig:latency_one_inference_step}
    \end{subfigure}
    \vspace{-0.5em}
    \caption{Latency comparison between parallel KV compaction and on-CPU multi-threaded memory management.}
    \vspace{-0.5em}
    \label{fig:latency_comparision_with_cpu_mem_management}
\end{figure}

\myparagraph{Memory Management Overhead}
Figure \ref{fig:latency_comparision_with_cpu_mem_management} compares Diff-KV's on-GPU parallel KV compaction with an alternative implementation that performs multi-threaded memory management on the CPU.
Evaluated with a sequence length of 1024 tokens across different batch sizes, DiffKV reduces memory management latency by up to three orders of magnitude relative to the on-CPU approach.
For one entire inference step, DiffKV significantly outperforms the on-CPU approach, particularly in the generation phase, where the memory management overhead on the CPU far exceeds the model execution time.
Therefore, \system's on-GPU parallel memory management is essential for ensuring that the performance benefits of KV cache compression are fully realized without being overshadowed by memory management overhead.

\begin{figure}[t]
    \centering
    \includegraphics[width=\linewidth]{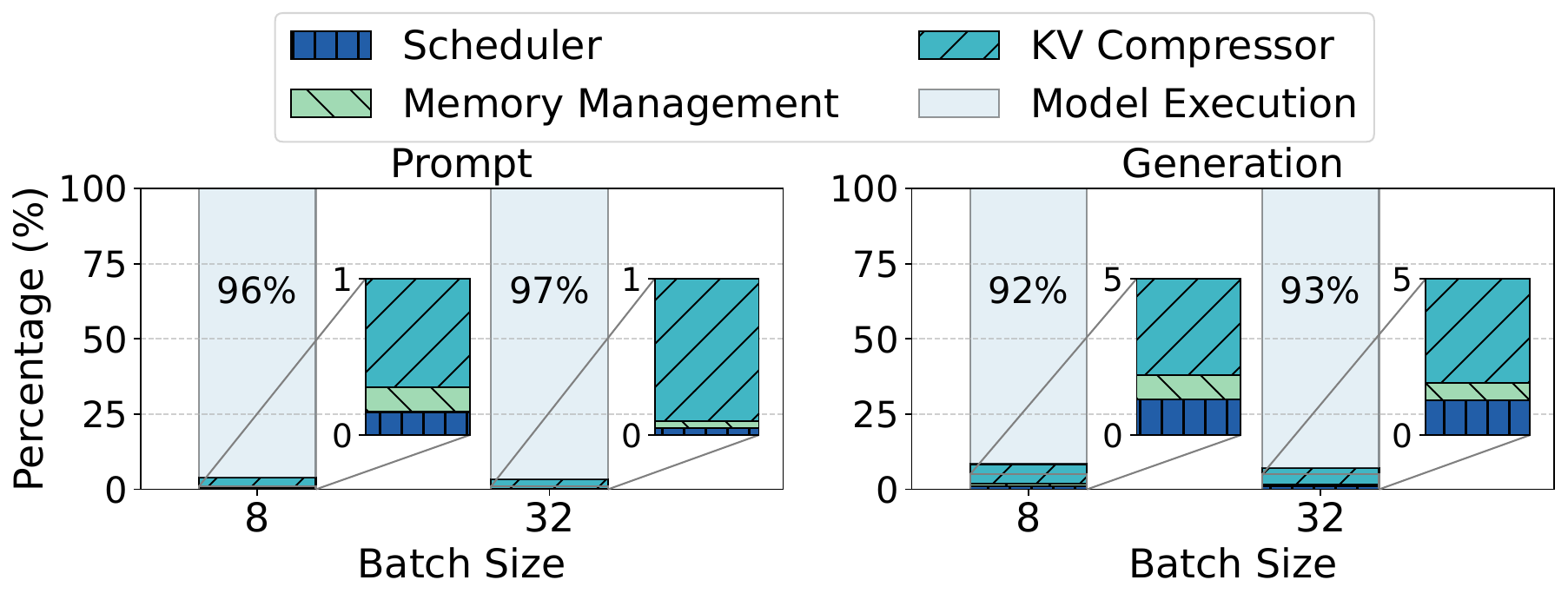}
    \vspace{-2em}
    \caption{Latency breakdown of \system.}
    \vspace{-1em}
    \label{fig:latency_breakdown}
\end{figure}

Figure \ref{fig:latency_breakdown} further illustrates the latency breakdown of Diff-KV during one inference step.
The memory management overhead is remarkably low thanks to \system's on-GPU parallel KV compaction, contributing less than 0.2\% of the total latency in the prompt phase and under 0.9\% in the generation phase.
Model execution dominates the latency in both phases, accounting for 96--97\% of the latency in the prompt phase and 92--93\% in the generation phase.
Notably, as the batch size increases, the percentage of time spent on model execution rises slightly, reflecting the scalability of \system.

\begin{figure}[t]
    \begin{subfigure}{0.49\linewidth}
        \centering
        \includegraphics[width=\linewidth]{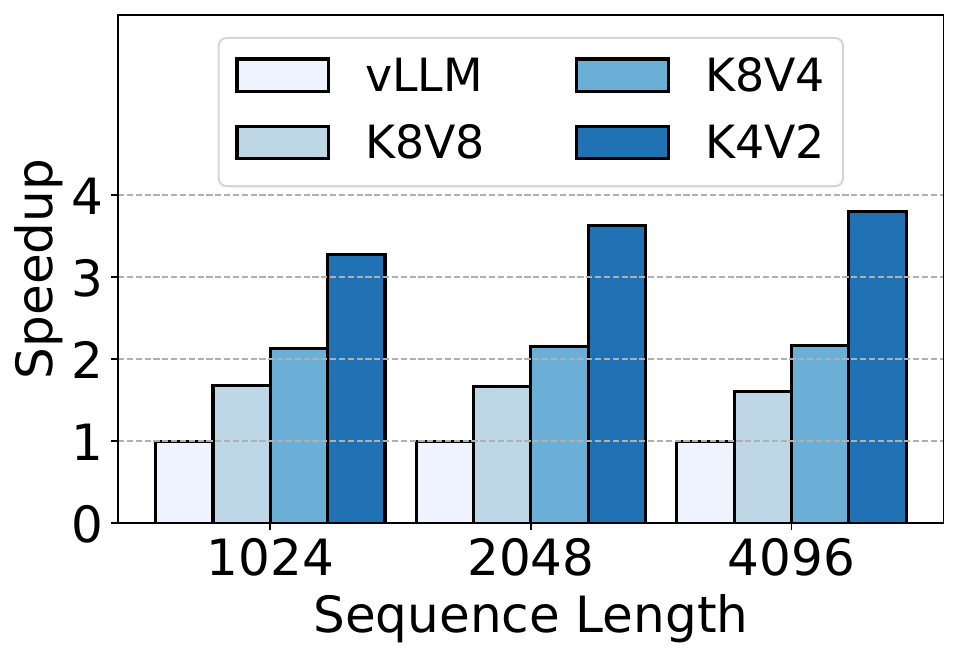}
        \vspace{-1.5em}
        \caption{Attention kernel}
        \label{fig:latency_attention_kernel}
    \end{subfigure}
    \hfill
    \centering
    \begin{subfigure}{0.49\linewidth}
        \centering
        \includegraphics[width=\linewidth]{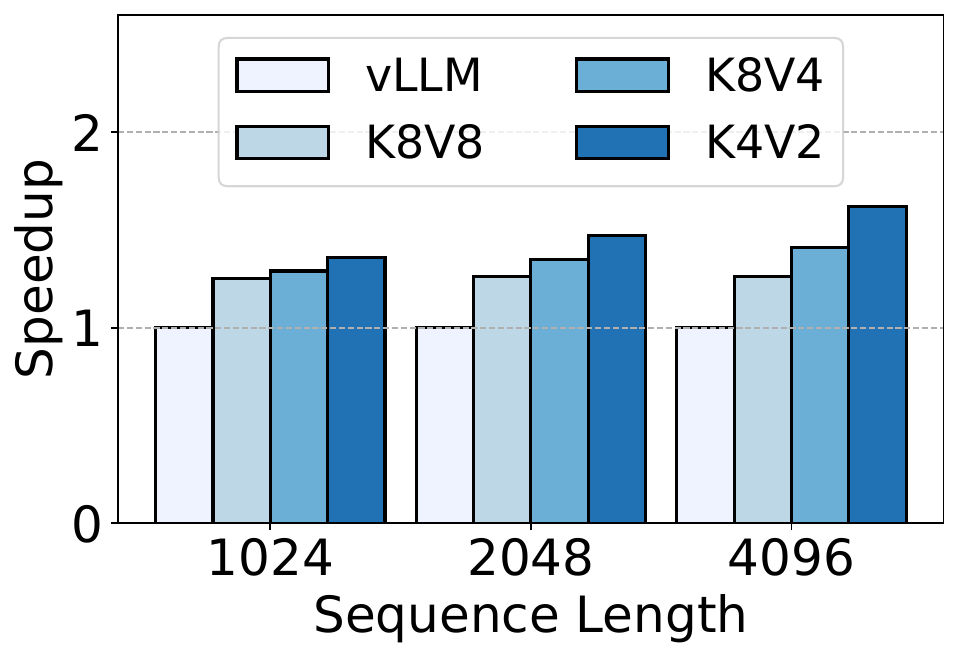}
        \vspace{-1.5em}
        \caption{One batch end-to-end}
        \label{fig:latency_one_batch}
    \end{subfigure}
    \vspace{-0.5em}
    \caption{Latency speedup of \system against vLLM.}
    \vspace{-1em}
    \label{fig:latency_speedup_over_vllm}
\end{figure}

\myparagraph{Latency Speedup}
Figure \ref{fig:latency_attention_kernel} shows the speedup of Diff-KV's custom attention kernel against vLLM under different quantization configurations.
\system achieves a near-linear speedup proportional to the reduction in KV cache size.
For example, with K8V8, which halves the KV cache size relative to FP16 datatype, the theoretical speedup is 2$\times$, and \system achieves 1.7$\times$.
The slight gap is primarily due to the overhead of accessing quantization metadata and performing dequantization.
Additionally, \system achieves greater speedups on longer sequences, indicating that its bandwidth optimization techniques become increasingly effective as the sequence length grows.
Figure \ref{fig:latency_one_batch} shows the end-to-end inference latency for a batch of 8 sequences.
With sequence length 4096, \system achieves a 1.4--1.6$\times$ speedup over vLLM.
Note that while \system does improve latency, its primary design focus is on enhancing throughput by supporting larger batch sizes.

\begin{figure}
\centering
\includegraphics[width=1\linewidth]{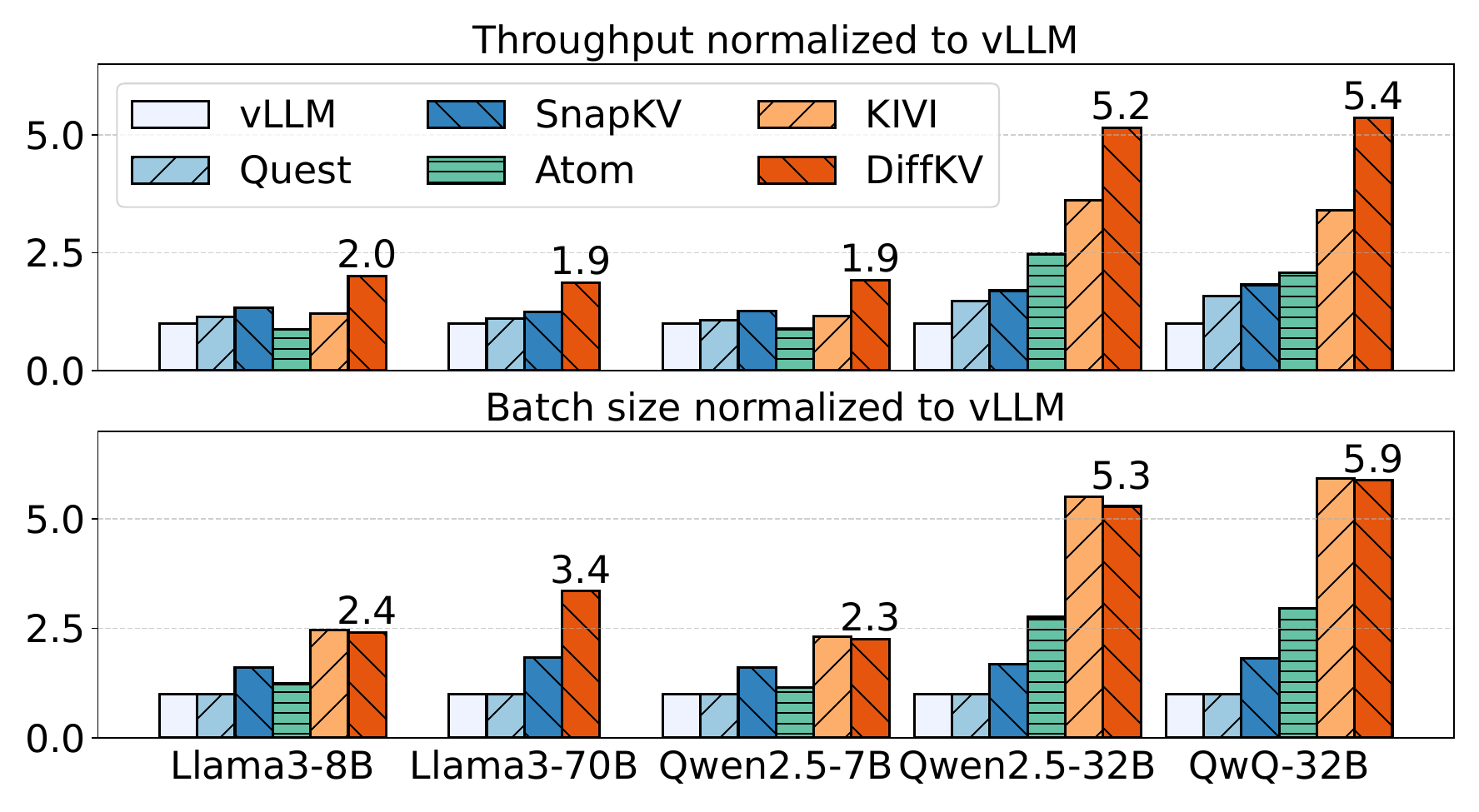}
\vspace{-2.em}
\caption{Throughput and achieved batch size.}
\vspace{-1em}
\label{fig:throughput}
\end{figure}

\myparagraph{Throughput Speedup}
Figure \ref{fig:throughput} presents the end-to-end throughput and achieved batch sizes of \system compared to vLLM, pruning-based approaches including Quest and SnapKV, as well as quantization-based methods including Atom (4-bit) and KIVI (2-bit).
Note that these pruning- and quantization-based baselines incur larger accuracy degradation than \system, as shown in Table~\ref{tab:accuracy_and_memory_usage}.
The maximum generation length is set to 16K tokens for QwQ-32B, 8K for Qwen2.5-32B, and 4K for the other models.
We evaluate on 1000 sequences sampled from the MATH dataset~\cite{lewkowycz2022solving}, which elicits chain-of-thought reasoning and typically leads to long generations reaching the specified limit.
Compression thresholds for \system are adopted from Figure~\ref{fig:thresh_calibration}.
Across all models, \system consistently achieves higher throughput than prior systems.
Notably, \system achieves a remarkable $5.4\times$ higher throughput over vLLM on the QwQ-32B thinking model, while Quest, SnapKV, Atom, and KIVI achieve $1.6\times$, $1.8\times$, $2.1\times$, and $3.4\times$ speedups, respectively.
This result highlights the superior efficiency of \system in processing long sequences, making it particularly well-suited for tasks that require extended reasoning.

The throughput improvements are directly correlated with the larger batch sizes supported by \system, enabled by its KV cache compression techniques.
For instance, with the QwQ-32B model, \system sustains a batch size of 15.9, far exceeding vLLM's batch size of 2.7.
The increased batch size enables \system to maximize the utilization of GPU compute capacity, leading to higher throughput.
In contrast, Quest supports the same batch size as vLLM because it retains the entire KV cache without reducing memory usage.
Its speedup comes primarily from faster attention computation, as it processes only a subset of tokens deemed important.
However, the process of estimating token importance incurs additional overhead, limiting the overall efficiency of Quest.
Note that although Atom and KIVI substantially increase the batch size, approaching or even exceeding \system, their throughput gains are not proportional.
This is because both methods are built on the HuggingFace Transformers library, which incurs higher framework overhead than vLLM and \system, and lacks high-performance GPU kernels that fuse dequantization with attention computation, as implemented in \system.
These results underscore that custom GPU kernels and an efficient runtime are essential for effectively translating KV cache compression into throughput gains.

\begin{figure}[t]
    \centering
    \includegraphics[width=0.95\linewidth]{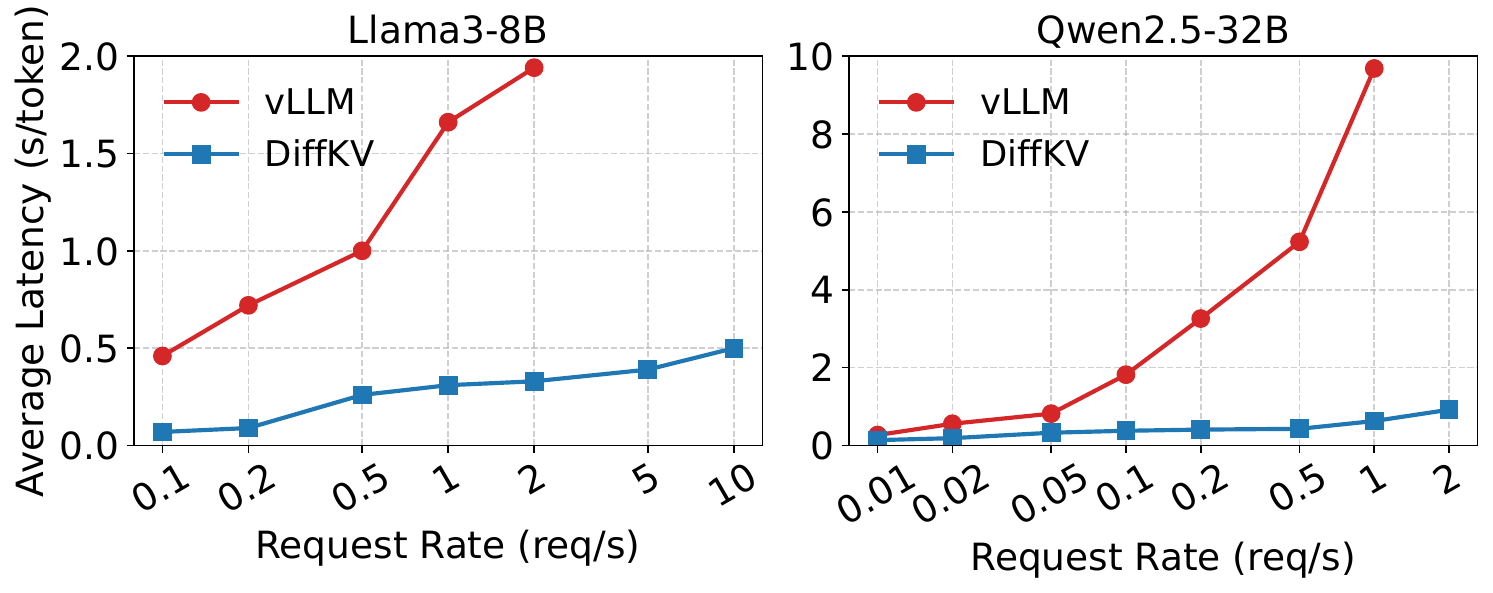}
    \vspace{-1em}
    \caption{Comparison of average latency between DiffKV and vLLM under dynamic workloads.}
    \vspace{-1em}
    \label{fig:eval_serving}
\end{figure}

\myparagraph{Serving Dynamic Workloads}
We evaluate \system under dynamic workloads and compare it to vLLM~\cite{kwon2023efficient}.
Following the methodology of vLLM, we inject requests according to a Poisson arrival process and sweep the request rate.
Figure~\ref{fig:eval_serving} reports the average per-token latency (including both queuing and LLM processing time) for Llama3‑8B and Qwen2.5‑32B across a range of request rates.
\system consistently achieves lower latency and sustains higher loads before queuing delays grow sharply.
These improvements stem from \system's reduced KV cache footprint, which enables larger batch sizes, and its parallel on-GPU KV compaction, which keeps memory management overhead low even under bursty traffic.

\vspace{-0.05in}

\section{Conclusion}

\vspace{-0.08in}

We proposed \system, which enhances LLM serving efficiency by exploiting three levels of differentiation in KV cache: differentiated precision for keys and values, hierarchical compression based on token importance, and per-head dynamic sparsity.
At the core of \system is the parallel KV compaction technique that efficiently handles irregular memory requirements across requests and attention heads, effectively translating memory savings into performance gains.
Our evaluation demonstrated that \system is able to compress the KV cache by $2.7\times$ to $5.7\times$ with near-lossless accuracy, even on thinking models and complex workloads that require sophisticated reasoning and long-generation capabilities, and enhances throughput by $1.9\times$ to $5.4\times$, outperforming prior KV cache compression methods.

\section{Acknowledgment}
We sincerely thank our shepherd, Tim Harris, for his thoughtful feedback and careful guidance throughout the paper shepherding process. We are also thankful to Mingzhe Hao, Shuang Chen, other friends, and the anonymous reviewers for their insightful comments on earlier versions of this manuscript.


\bibliographystyle{acm}
\bibliography{references}
\end{document}